%% file: main.tex
\documentclass[
    10pt,
    twocolumn,
    a4paper,
]{article}

\usepackage[utf8]{inputenc}
\usepackage[
    colorlinks=true,
    linkcolor=black,
    citecolor=black,
    filecolor=black,
    urlcolor=black,
    plainpages=false,
    pdfpagelabels,
    breaklinks=true,
    pdftitle={DelGrad: Exact event-based gradients for training delays and weights on spiking neuromorphic hardware},
    pdfauthor={Julian Göltz, Jimmy Weber, Laura Kriener, Sebastian Billaudelle, Peter Lake, Johannes Schemmel, Melika Payvand, Mihai A. Petrovici},
]{hyperref}
\usepackage[
    style=nature,
    citestyle=numeric-comp,
    sorting=none,
    giveninits=true,
    uniquename=init,
    maxbibnames=3,
    minbibnames=2,
    isbn=false,
    doi=false,
    url=false,
]{biblatex}
\usepackage{amsthm,amsmath,amsfonts, amssymb, amscd}
\usepackage{bbm}
\usepackage{bbold}
\usepackage{booktabs}
\usepackage{caption}
\usepackage[capitalize, nameinlink, poorman]{cleveref}
\usepackage{wrapfig}  %
\usepackage[]{geometry} %
\usepackage[docdef=atom]{glossaries-extra}
    \setabbreviationstyle[acronym]{long-short}
    \glssetcategoryattribute{acronym}{nohyperfirst}{true}

    \input{glossaries.tex}
\usepackage{graphicx} %
\newcommand*{\linelabel}[1]{}
\usepackage{mathtools} %
\usepackage{placeins}  %
\usepackage{threeparttable} %
\usepackage{slashed}  %
\usepackage{siunitx}
\usepackage{tikz}
    \usepackage{pgfplots}
    \pgfplotsset{compat=1.18}
    \usetikzlibrary{calc}
\AddToHook{cmd/section/before}{}%
\newcommand*\subtxt[1]{_{\textnormal{#1}}}
\DeclareRobustCommand\_{\ifmmode\expandafter\subtxt\else\textunderscore\fi}
\mathchardef\ordinarycolon\mathcode`\:
    \mathcode`\:=\string"8000
    \begingroup \catcode`\:=\active
    \gdef:{\mathrel{\mathop\ordinarycolon}}
    \endgroup
\newcommand{\LTinput}[1]{}
    \newcommand{\YYCleverefInput}[1]{}
    \LTinput{main.glsdefs}
    \YYCleverefInput{main.sed}
\newcommand{\vect}[1]{\mathbf{#1}}
\DeclareMathOperator*{\argmax}{arg\,max}

\newcommand{\gl}{\gL}\newcommand{\gL}{g_\ell}
\newcommand{\intd}{\textrm{d}}
\newcommand{\Is}{I\_{s}}
\newcommand{\taum}{\tau\_{m}}
\newcommand{\tauref}{\tau\_{ref}}
\newcommand{\taurefone}{\tau\_{ref,1}}
\newcommand{\taureftwo}{\tau\_{ref,2}}
\newcommand{\taus}{\tau\_{s}}
\newcommand{\tdel}{t^{d}}
\newcommand{\tnodel}{t^{\slashed d}}
\newcommand{\T}{T}

\newcommand{\um}{u\_{m}}
\newcommand{\umdot}{\dot u\_{m}}
\newcommand{\Vleak}{E_\ell}
\newcommand{\Vreset}{V_\text{reset}}
\newcommand{\Vth}{\vartheta}
\newcommand\shorthandU{u({\scriptstyle t | \{\theta\}})}
\newcommand\shorthandUi[1]{u_{#1}({\scriptstyle t | \{\theta\}})}
\newcommand\shorthandIntegrationDomain{S}
\title{
DelGrad: Exact event-based gradients
\\
for training delays and weights
\\
on spiking neuromorphic hardware
}

\newcommand{\affilKIP}{1}
\newcommand{\affilPhysio}{2}
\newcommand{\affilINI}{3}
\newcommand{\sharedFirst}{*}
\newcommand{\sharedSenior}{\S}
\author{
    \\[0em]
    \normalsize{
        Julian Göltz\textsuperscript{\sharedFirst,\,\affilKIP,\,\affilPhysio},
        Jimmy Weber\textsuperscript{\sharedFirst,\,\affilINI},
        Laura Kriener\textsuperscript{\sharedFirst,\,\affilINI,\,\affilPhysio},
    }
    \\
    \normalsize{
        Sebastian Billaudelle\textsuperscript{\affilINI,\,\affilKIP},
        Peter Lake\textsuperscript{\affilKIP},
        Johannes Schemmel\textsuperscript{\affilKIP},
    }
    \\
    \normalsize{
        Melika Payvand\textsuperscript{\sharedSenior,\,\affilINI},
        Mihai A. Petrovici\textsuperscript{\sharedSenior,\,\affilPhysio}
    }
    \\[0.5cm]
	{\normalfont\small
        \textsuperscript{\sharedFirst} shared first authorship
        \hspace{1cm}
        \textsuperscript{\sharedSenior} shared senior authorship
    }
    \\[0.1cm]
	{\normalfont\small
         \textsuperscript{\affilKIP}\,Kirchhoff-Institute for Physics, Heidelberg University.
    }\\{\normalfont\small
        \textsuperscript{\affilPhysio}\,Department of Physiology, University of Bern.
    }\\{\normalfont\small
        \textsuperscript{\affilINI}\,Institute of Neuroinformatics, University of Zurich and ETH Zurich.
    }
}
\date{}

\begin{document}
\maketitle

\glsresetall

\begin{abstract}
\Glspl{snn} inherently rely on the timing of signals for representing and processing information.
Incorporating trainable transmission delays, alongside synaptic weights, is crucial for shaping these temporal dynamics.
While recent methods have shown the benefits of training delays and weights in terms of accuracy and memory efficiency, they rely on discrete time, approximate gradients, and full access to internal variables like membrane potentials.
This limits their precision, efficiency, and suitability for neuromorphic hardware due to increased memory requirements and I/O bandwidth demands.

To address these challenges, we propose DelGrad, an analytical, event-based method to compute exact loss gradients for both synaptic weights and delays.
The inclusion of delays in the training process emerges naturally within our proposed formalism, enriching the model's search space with a temporal dimension. 
Moreover, DelGrad, grounded purely in spike timing, eliminates the need to track additional variables such as membrane potentials. 
To showcase this key advantage, we demonstrate the functionality and benefits of DelGrad on the BrainScaleS-2 neuromorphic platform, by training \glspl{snn} in a chip-in-the-loop fashion. 
For the first time, we experimentally demonstrate the memory efficiency and accuracy benefits of adding delays to \glspl{snn} on noisy mixed-signal hardware. 
Additionally, these experiments also reveal the potential of delays for stabilizing networks against noise.
DelGrad opens a new way for training \glspl{snn} with delays on neuromorphic hardware, which results in fewer required parameters, higher accuracy and ease of hardware training.
\end{abstract}

\section{Introduction}

\begin{figure*}[ht]
    \centering
    \includegraphics{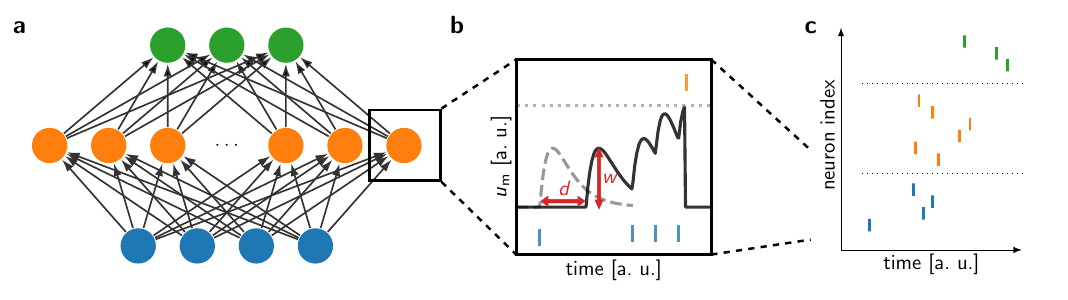}
    \caption{%
        \label{fig:intuition}
        \textbf{Information flow in a \glsxtrfull{snn}.}
        \textbf{a)} Network architecture of a feed-forward \gls{snn} with a spiking input layer at the bottom, a hidden layer in the middle and the output layer on top.
        While the methods described in this manuscript are applicable to many different network architectures, the structure depicted in a), with variable size of the hidden layer, is used in the following.
        \textbf{b)} Zoom-in on the information processing in a single \glsxtrfull{lif} neuron in the hidden layer.
        Incoming spikes (blue, bottom) are integrated by the neuron's membrane $u_{m}$ and generate \glspl{psp}, which accumulate additively.
        Once the membrane potential passes a threshold (gray dashed line), an output spike (orange, top) is generated and passed on to the neurons in the next layer. 
        The \gls{psp} amplitudes are modulated by the respective synaptic weights $w$ (vertical red arrow); these are the parameters that are conventionally adapted during learning.
        Learnable transmission delays $d$ (horizontal red arrow) shift \glspl{psp} in time, providing additional temporal processing power to the neuron.
        \textbf{c)} Zoom-out to a raster plot of the full spiking activity in the network.
        The information passed between the layers is encoded in the timing of the spikes.
        As sketched in the raster plot, in the experiments in this manuscript we employ \glsxtrshort{ttfs} coding, i.e., each neuron spikes only once, however our method also generalizes to multi-spike scenarios (\cref{sec:si_math_multi}) if required by the task.
    }
\end{figure*}

The mammalian brain has always represented the ultimate example of computational prowess, and therefore remains an important source of inspiration for understanding intelligence and replicating it in artificial substrates.
In particular, its specific mechanisms for transmitting and processing information have been the subject of intense scrutiny and debate.
Among these, the pulsed communication between neurons, predominantly based on all-or-none events, called action potentials or spikes, stands out as a distinguishing feature, and has thus been suggested to play an important role in the brain's remarkable combination of computational performance and energy efficiency~\cite{olshausen1996emergence, koch2000role}.
Consequently, spike-based communication represents a de facto standard across current neuromorphic platforms, which aim to inherit the proficiency of their biological archetype by replicating chosen aspects of its structure and dynamics~\cite{mead1990neuromorphic, indiveri2015memory, frenkel2023bottom, furber2014spinnaker, billaudelle2020versatile}.

Among the various encoding schemes proposed for spiking neurons, the representation of information within the specific timing of individual spikes is of particular interest~\cite{bohte2004evidence}, as it effectively allows the communication of, ideally, real-valued signals on an energy budget equivalent to only generating and transmitting a single bit.
This gives rise to a specific call for \gls{snn} training algorithms that exploit the temporal richness of spike timing codes for solving computational tasks efficiently and accurately, while remaining capable of operating under the realistic constraints of the underlying physical substrate, whether biological or artificial. 

Recent years have seen an exciting trend in this direction, showing how the performance of \glspl{snn} can be improved by optimizing various temporal parameters.
Such parameters include neuronal integration time constants~\cite{yin2021accurate, rao_etal2022_sLSTM,nowotny2022loss, bittar2022surrogate, Perez_Nieves_2021,fang_etal2021_timeconstant, Moro2024hierarchy}, adaptation time constants~\cite{bellec_etal2018_lsnn}, and delay variables~\cite{hammouamri_etal2022_threshold,dagostino_etal2024_denram,shrestha2018SLAYER}.

In particular, spike transmission delays have been predicted to significantly enrich the information processing capabilities of spiking networks~\cite{maass1999complexity, Izhikevich2006}, but specific applications to computationally demanding tasks had remained an open issue.
However, recent evidence suggests that a co-optimization of synaptic weights and delays is possible and can significantly reduce the number of training parameters in an \gls{snn}, without loss of accuracy~\cite{hammouamri2023learning, patino_etal2023_imec_delay}.
This finding is especially important for neuromorphic architectures that target limited on-chip memory.

Nevertheless, from an \emph{algorithmic} perspective, optimizing delays in \glspl{snn} remains an ongoing research problem.
\linelabel{rev:refCitationOfDelaySelection}
Previous literature has largely focused on either exploiting heterogeneity in delay parameters, while limiting gradient-based training to the weights, resulting in selecting the suitable delays~\cite{patino_etal2023_imec_delay,dagostino_etal2024_denram,Bohte_2002,Gerstner1996delays},
or using evolutionary, not gradient-based algorithms to find delay parameters~\cite{schuman_etal2020_eons}.
Recently, several approaches based on surrogate gradients~\cite{neftci2019surrogate} have been proposed, using convolutional kernels~\cite{hammouamri2023learning} or finite difference methods~\cite{shrestha2018SLAYER,sun_etal2023_axonal}.
\linelabel{rev:refSurrogateDesc}
The underlying surrogate-gradient approach partially addresses, at the expense of both exactness and the conservative property the gradient field~\cite{gygax2025surrogateGradients}, the  discontinuities of (dis)appearing spikes by smoothing out the spiking threshold.
These types of algorithms operate in discrete time, which requires the storage of neuronal activities as binary vectors over the entire history of the \gls{snn}.

In addition, from a \emph{hardware} perspective, there is a growing number of neuromorphic platforms that support the emulation of delays.
These implementations require additional memory elements and parameter sets to retain the information of the incoming spike for a controllable amount of time.
Previous implementations of on-chip delays using \gls{cmos} technology have used digital circuits~\cite{madhavan_etal2014_racelogic,davies2018loihi,patino_etal2023_imec_delay,madhavan2021temporal,merolla2014_truenorth}, active analog circuits~\cite{sheik_etal2012_analogdelay,wang_etal2011_andredelay,Huayaney_etal16,gerber_etal2022_dynap1}, or mixed-signal solutions~\cite{Richter_etal24}. 
Furthermore, emerging memory technologies such as \gls{rram} have also been used to realize delay elements, taking advantage of their non-volatile, small three-dimensional footprint, and zero-static-power properties~\cite{dagostino_etal2024_denram,madhavan2021temporal}.
This increasing abundance of neuromorphic substrates offering configurable delays reveals an implicit call for algorithms capable of exploiting these novel capabilities.

In this work, we present DelGrad, which to the best of our knowledge is the first exact, analytical solution for gradient-based, hardware-compatible co-learning of delays and weights, using exclusively spike times for the computation of parameter updates.
\linelabel{rev:refCodesign}
Its hardware compatibility stems from an algorithm-hardware co-design approach: the algorithm was developed with physical hardware systems in mind.
It considers the real-valued nature of spike times on analog systems, as well as system-level constraints such as low I/O bandwidth and limited on-chip memory, resulting in a method that is inherently hardware-friendly.
Compared to previous approaches, this spike-time-based method simultaneously increases precision and computational efficiency, while also minimizing the required memory footprint of the model.
Under DelGrad, we quantitatively study the effect of different types of delays in relation to the size and performance of \glspl{snn}.
Finally, to experimentally demonstrate the efficacy of our approach, we utilize DelGrad to perform chip-in-the-loop training of a delay-based \gls{snn} on a mixed-signal neuromorphic chip.

\section[Training delays in SNNs with DelGrad]{Training delays in \glspl{snn} with DelGrad}\label{sec:theory}

While \glsfmtlongpl{snn} share their overall structure with the more well-known \glsfmtlongpl{ann}, the intrinsic dynamics in the networks are different.
In \glspl{snn}, each unit in the network is a model of a \emph{spiking} neuron, i.e., communicating with binary all-or-nothing events called spikes, that carry information in their precise timing (\cref{fig:intuition}).
Here, we employ the \gls{lif} neuron model, which despite its relative simplicity captures the most important properties of biological neurons and therefore often serves as a ``standard model'' in computational neuroscience and neuromorphic engineering~\cite{lapicque1907recherches,Abbott_1999}.
Among these properties are foremost the spiking communication
and the leaky-integrator dynamics: all input events are integrated on the membrane potential which, after an excitation, slowly decays back to its resting state.
The precise dynamics of a network of \gls{lif} neurons are determined by parameters of the neurons, but also by the inter-neuron connectivity and its parametrization which includes synaptic weights and transmission delays.

To train the parameters of an \gls{ann}, the error backpropagation algorithm~\cite{linnainmaa1970representation,Rumelhart1986}, which optimizes parameters via gradient descent, has become the de facto standard.
In contrast, only recently has it become clear that in the case of \glspl{snn} the non-differentiability of spikes is, in fact, not an impediment for performing gradient-based optimizations.
\linelabel{rev:refCitationStan}
The developed optimization methods for training \glspl{snn} can be roughly split into two groups: approximate, surrogate gradient approaches~\cite{neftci2019surrogate,Zenke_2018_superspike,renner2024backpropagation,shrestha2018SLAYER}, and methods that employ exact spike time gradients~\cite{goeltz2021fast,wunderlich2021eventprop,Bohte_2002,Klos_2025_QIFsnn}.
In the following, we base our study on the exact gradient methods described in~\cite{goeltz2021fast}.

\paragraph{Spike time gradients}

The subthreshold dynamics of the membrane potential $\um$ of an \gls{lif} neuron with exponential current-based synapses
is governed by the differential equation
\begin{equation} \label{eq:cubaLIF}
	\taum \umdot(t) = [\Vleak - \um(t)] + \Is(t) / \gl 
\end{equation}
with neuronal time constant $\taum$, leak potential $\Vleak$, leak conductance $\gl$ and synaptic input current $\Is$, defined as 
\begin{equation} \label{eq:Is}
\Is(t) = \sum_i \Theta(t-t_i) w_i \exp(-(t-t_i)/\taus)
\end{equation}
where $\Theta(t-t_i)$ is the Heaviside step
function, 
$w_i$ is the weight associated with the synapse receiving a
spike at time $t_i$, and $\taus$ is the synaptic time constant.
 $\Is$ thus outputs a current which is a first-order low pass filter of the input spike train, represented by the exponential kernel $\exp(-(t-t_i)/\taus)$.
$\Is$ is itself further leaky integrated by the neuron's membrane, with time constant $\taum$.
Upon crossing the spiking threshold $\Vth$, the membrane is reset to $\Vreset$ for a refractory period $\tauref$, during which the neuron does not react to any further input spikes, and the neuron emits an output spike.

The response function of a neuron thus, fundamentally, maps a sequence of input spike times $t_i$ to a sequence of output spike times $T_i$.
\linelabel{rev:multiSpikeReference}
    Here, we focus on a single output spike per neuron for ease-of-notation but the method can be straightforwardly extended to multi-spike scenarios, as described in~\cref{sec:si_math_multi}.
For one such output spike time $T$, under a parametrization given by the synaptic weights $w_i$ of the incoming connections, we can write $T$ as a function of the set of input weights $\{w_i\}$ and set of input spikes $\{t_i\}$:

\begin{equation}
    \label{eq:T}
    \T\left(\{t_i\} \cup \{w_i\}\right) \; .
\end{equation}
Under certain conditions, depending on the values of the neuronal and synaptic time constants $\taum$ and $\taus$, the function $T$ becomes analytic, as discussed in \cite{goeltz2021fast}.

For example, for $\taum=\taus$ 

\begin{equation}
    \label{eq:equalTimeEquation}
    \T  =  \taus \left\{
                \frac{b}{a_1} - \mathcal{W}\!\left[
                       -\frac{\gL\Vth}{a_1} \exp\left(\frac{b}{a_1}\right)
                \right]
            \right\}
        \; ,
\end{equation}
and for $\taum=2\taus$ 
\begin{equation}
    \label{eq:doubleTimeEquation}
    \T  =  2\taus\ln\! \left[
                \frac{2a_1}{a_2 + \sqrt{a_2^2 - 4a_1\gL\Vth}}
            \right]
        \; ,
\end{equation}
where $a_i$ and $b$ are explicit functions of $w_i,t_i$ and $\mathcal{W}$ is the Lambert W function (see~\cref{eq:SI_mathAB}).
Writing the spike time in this way enables us both to perform an efficient, event-based forward pass and to train this network, by calculating the exact gradient of the output of the network with respect to the network parameters.

\linelabel{rev:multiSpikeReferenceTwo}
In a multi-spike scenario (\cref{sec:si_math_multi}), all subsequent spikes after the first spike can be calculated by taking the reset into account and solving the equation for different initial conditions.
Ultimately, this results in similar expressions as~\cref{eq:T,eq:equalTimeEquation,eq:doubleTimeEquation}.

To optimize the network parameters via gradient descent, we base the update of each parameter $\theta$ on its influence on the loss $\mathcal{L}$, the gradient $\partial\mathcal{L} / \partial \theta$.
Employing the chain rule, this gradient is iteratively composed of $\partial T / \partial \theta$ and ${\partial \T}/{\partial t_i}$, i.e., derivatives of the above equations.

Specifically for a network with parameters $w_i$, ${\partial \T}/{\partial w_i}$ allows us to link a deviation in an output spike time to a change in weight parameters, while ${\partial \T}/{\partial t_i}$ relates this deviation in output to deviation in the input, thereby enabling us to propagate an error in the spike time backwards through the neuron.
Crucially, just like the original~\cref{eq:doubleTimeEquation,eq:equalTimeEquation}, and in contrast to surrogate-gradient-based approaches, these derivatives only depend on spike times and parameters of the network and can be computed without the calculation or measurement of the membrane potential.
This allows us to perform a fully event-based forward and backward pass, without any need for temporal discretization of forward or backwards dynamics.

Transmission delays of spike signals can now simply be introduced as additive parameters $d$ to the original spike times $\tnodel$:
\begin{align}
    \label{eq:delayIsAddition}
    \tdel_i &= \tnodel_i + d_i \; .
\end{align}
These delayed spike times then become the relevant input for the postsynaptic neuron.
As above, derivatives of this expression provide the necessary quantities for adapting the delays and for backpropagating the spike timing errors.
In this case, the corresponding equations are trivial:
\begin{align}
    \label{eq:delay_derivatives}
    \frac{\partial \tdel_i}{\partial \tnodel_i} = 1
    \quad
    \text{and}
    \quad
    \frac{\partial \tdel_i}{\partial d_i} = 1
    \; .
\end{align}

Treating spike times as continuous variables, different from the time-binning performed in other approaches~\cite{shrestha2018SLAYER,hammouamri2023learning}, allows this natural implementation of full-precision delays as well as the exact and simple training of the delay parameters.
We note that these considerations do not depend on a specific network setup and thus apply to any activity pattern in arbitrary spiking networks.
In the following, we focus our attention on the particular problem of pattern classification, for which we employ a specific network architecture and spike coding scheme (\cref{fig:intuition}).

\begin{figure*}[!t]
    \centering
    \begin{tikzpicture}[inner sep=0pt, anchor=north west, scale=0.9]
        \draw[use as bounding box, draw=none] (0, 0) rectangle (7.12in, -3.5in);
        \node (panelLabelA) at (0.2, -0.1) {\large\sffamily\textbf{a}$\vphantom{\int}$};
        \node (panelLabelB) at (1.6in, -0.1) {\large\sffamily\textbf{b}$\vphantom{\int}$};
        \node (panelLabelC) at (4.4in, -0.1) {\large\sffamily\textbf{c}$\vphantom{\int}$};
        \node (panelA) at ([xshift=-0.2cm, yshift=-0.15cm]panelLabelA) {
            \includegraphics[scale=0.9]{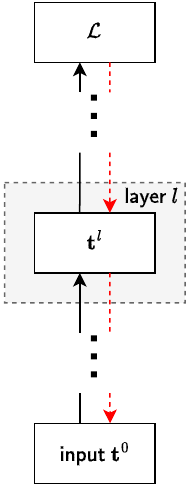}};
        \node (panelB) at ([xshift=0.3cm, yshift=-0.05cm]panelLabelB) {
            \includegraphics[scale=0.9]{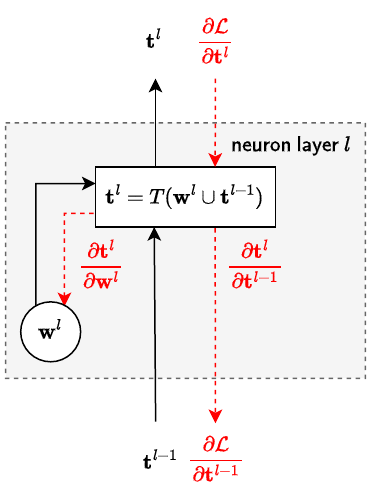}};
        \node (panelC) at ([xshift=0.5cm, yshift=-0.05cm]panelLabelC) {
            \includegraphics[scale=0.9]{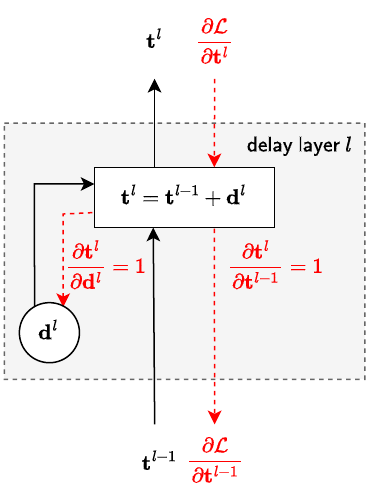}};
    \end{tikzpicture}
    \caption{
        \label{fig:computationalGraph}
        \textbf{Computational graph of a multi-layer \gls{snn} with spike-time information encoding and adjustable delay and weight parameters.}
        \textbf{a)} Graph for a multi-layer network with spike times $\vect t^0$ injected into the bottom ($1^\text{st}$) layer.
        In the forward pass (black arrows), each layer $l$ takes spike times as inputs and returns spike times as outputs that go into the next layer.
        The spike times of the topmost layer are used to compute the loss function $\mathcal{L}$.
        The backward pass (red dashed arrows) starts at the loss and passes the gradients backwards through the layers.
        We consider two types of layer: neuron layers and delay layers.
        \textbf{b)} Neuron layer with parameters $\vect w^l$ (synaptic weights).
        These are used together with the input spike times $\vect t^{l-1}$ to calculate the output spike times $\vect t^l$ according to the nonlinear relation described in \cref{eq:equalTimeEquation,eq:doubleTimeEquation}.
        \textbf{c)} Delay layer with parameters $\vect d^l$ that are added (linearly) to the input spike times $\vect t^{l-1}$ to calculate the output spike times $\vect t^l$ as in \cref{eq:delayIsAddition}.
    }
\end{figure*}

\paragraph{Extension to a multi-layer network}
To take advantage of a well-established architectural paradigm, we now consider information propagation in hierarchical feed-forward networks.
As also shown in the corresponding computational graph (\cref{fig:computationalGraph}a, solid black arrow), the input
$\vect t^0$ is passed through the sequence of layers until it reaches the output.\footnote{We use bold symbols to denote non-scalar variables.}
The gradient of the chosen loss function $\mathcal{L}$ then goes backwards through the network (dashed red arrow) for optimizing the parameters.
In the forward pass, the only information that is transmitted are spike times $\vect t^l$; in the backward pass, we transmit the gradient of the loss function ${\partial \mathcal{L}}/{\partial \vect{t}^l}$, but note that it is also only evaluated at the times when neurons spike.

\linelabel{rev:refVirtualLayer}
For \glspl{snn} with delays, the computational graph differentiates between two types of layer: neuron layers and delay layers (\cref{fig:computationalGraph}).
Both layers receive input spikes $\vect{t}^{l-1}$  and return output spikes $\vect{t}^l$, but using different forward transfer functions, as given by \cref{eq:equalTimeEquation}/\cref{eq:doubleTimeEquation} and \cref{eq:delayIsAddition}, respectively.
In the backward direction, they pass the partial derivative ${\partial \mathcal{L}} / {\partial \mathbf{t}^{l-1}}$ discussed above.

\Cref{fig:computationalGraph}b and c highlight the similarity of the two layer types: both neuron and delay layers take spike trains as an input and produce spike trains as an output in the forward pass, and propagate gradients of the loss with respect to the corresponding spike times in the backward pass.
Their respective computations are carried out sequentially, as depicted in \cref{fig:computationalGraph}a, with delay layers stacked in between neuron layers.

\begin{figure*}[ht]
    \centering
    \includegraphics{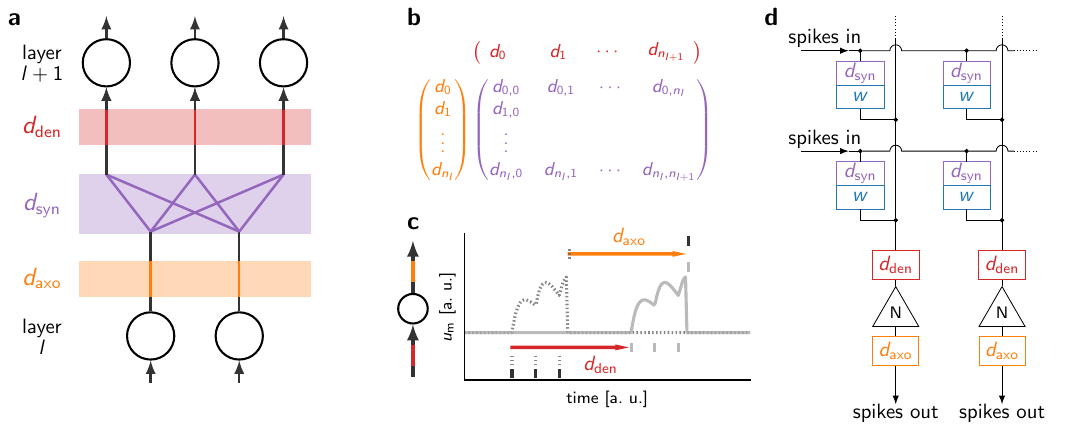}
    \caption{%
        \label{fig:delayConcepts}
        \textbf{Illustrating different types of delays.}
        \textbf{a)} From bottom to top: axonal delays shift the timing of the neuron's outgoing spikes by $d_\text{axo}$ (orange);
        synaptic delays shift the timing of spikes by a specific value $d_\text{syn}$ for each pair of pre- and post-synaptic neuron (purple); dendritic delays shift the timing of the  incoming spikes into a neuron by $d_\text{den}$ (red).
        \textbf{b)} Vector and matrix representation of the different types of delays and their dimensionality as a function of the number of pre- and post-synaptic neurons.
        \textbf{c)} Equivalent effect of the dendritic and axonal delays on the output spike time of a neuron, due to the time-shift invariance of the temporal dynamics of a \gls{lif} neuron.
        \textbf{d)} Schematic illustration of the location of synaptic, dendritic and axonal delay components in a generic neuromorphic crossbar architecture.
    }
\end{figure*}

\paragraph{Delay implementation}
In \cref{fig:delayConcepts}a we distinguish between different types of delays:
axonal delays $d_\text{axo}$ on a neuron's output, dendritic delays $d_\text{den}$ on a neuron's input, and synaptic delays $d_\text{syn}$ that are specific for every connection between pairs of neurons.
Their respective natural representations as column vectors, row vectors and matrices are shown in \cref{fig:delayConcepts}b.
The memory footprint of axonal and dendritic delays thus scales linearly with the number of neurons in the network, while for synaptic delays, it scales linearly with the network depth and quadratically with its width.

While in principle different types of delays can be simultaneously present in a network and can be combined with each other, it is important to note that, as illustrated in \cref{fig:delayConcepts}c, combining dendritic and axonal delays for the same neuron is redundant:
as neuronal dynamics are invariant to temporal shifts, it is equivalent for the input spikes to arrive with a delay
$d_\text{den} = d$, resulting in a delayed output spike (red arrow and gray curve), or for the output spike of the neuron to be directly delayed with $d_\text{axo} = d$ (orange arrow and membrane dynamics in black).

Given the resource constraints of neuromorphic systems, we investigate the performance benefits incurred by the different delay types, with different requirements on the memory resources.
Although a quantitative evaluation of the exact
energy consumption, chip area and design complexity of different delay architectures heavily depends on system architecture and the chosen design (e.g., analog vs.\ digital and circuit topology), some generic statements can be made using the mathematical representation of the delay elements.

For typical crossbar architectures (\cref{fig:delayConcepts}d), the synaptic delay mechanisms are often located within the crossbar array and therefore scale with the product of array's input and output size. 
In contrast, dendritic and axonal delays can be located in the periphery of the array, and thus their required area scales linearly with the input and output array size, respectively.
It is worth noting that an important property of axonal delay mechanisms is that they are located directly after the neurons' output and therefore only need to operate on sparse events.
In contrast, dendritic delays are located directly before the neurons' input, and after the input signals have been scaled by the synaptic weight.

Depending on the design choices, in particular on whether the synaptic integration happens in the synapses or in the neurons, this may require more complex circuitry. 
Note also that neurons usually receive more spikes than they emit, so the required buffering may also increase the corresponding hardware footprint of dendritic delay implementations.

\section{Simulation results}\label{sec:sim_results}

\begin{figure*}[ht]
    \centering
    \includegraphics{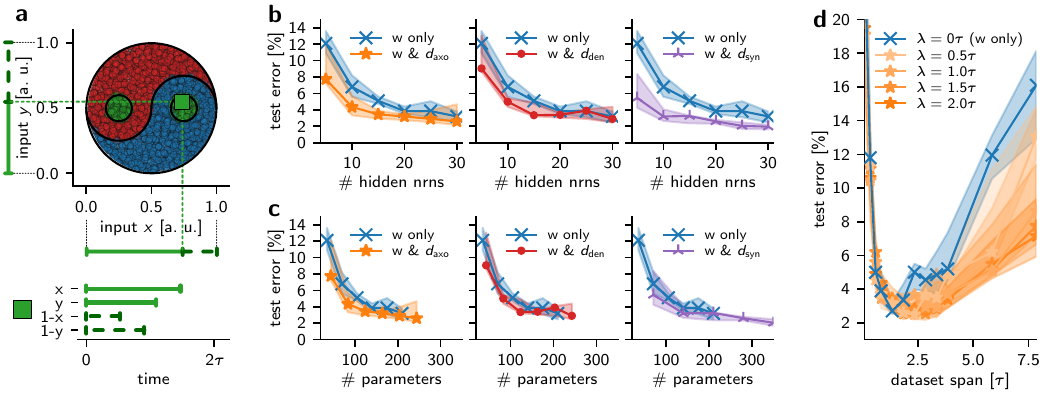}
    \caption{%
        \label{fig:simulation}
        \textbf{Classification task and simulation results.}
        \textbf{a)}
        The \glsxtrfull{yy} task \cite{kriener2021yin} consists of the classification of dots based on whether they belong to the Yin (red), Yang (blue), or dot (green) regions, as illustrated in \ref{fig:simulation}a.
        The input features are the two-dimensional coordinates $(x, y)$ of the image, along with their mirrored values $(1-x, 1-y)$, totaling four features.
        These features are encoded into spike times, such that a larger value of $x$ or $y$ coordinate results in a later spike time for $x$ or $y$  and an early spike time for its mirrored version $1-x$ or $1-y$ respectively.
        For more details on the encoding, see the original publication \cite{kriener2021yin}.
        \textbf{b)} Test error as a function of the number of hidden neurons in an \gls{snn}, using different delay types.
        The solid lines and markers show the median of the error, and the shaded areas illustrate the \glspl{iqr} for $10$ seeds.
        \textbf{c)} Same data as in b) but as a function of the number of trainable parameters in the networks, i.e., counting the distinct weights and, if applicable, delays.
        \textbf{d)} Impact of axonal delays as a function of the temporal scale of the dataset.
        The trainable delays cover a range $\lambda$ as indicated by the orange hue.
        The network performance without delays is shown in blue.
    }
\end{figure*}

This section evaluates DelGrad’s ability to co-train delays and weights, demonstrating improved accuracy and parameter efficiency over weight-only training. By systematically studying the effect of hidden layer size and comparing different delay types, we highlight the advantages of incorporating learnable delays.

\paragraph{Setup}
We benchmark a PyTorch \cite{paszke2019pytorch} implementation of the DelGrad method using the \gls{yy} \cite{kriener2021yin} dataset, to evaluate the impact of transmission delays on the \gls{snn} performance, and assess how this varies with the network size.
\linelabel{rev:refYYdifficult}
This dataset is selected for its advantageous properties -- compactness, making it amenable for hardware prototyping, training speed, as well as discriminatory power between network architectures and training paradigms:
it leaves ample room for benchmarking above the accuracy achievable with a linear classifier.
The task is to classify the region of a Yin-Yang image to which a point in the image plane belongs,
as illustrated in Fig.\ref{fig:simulation}a.
The coordinates of the point $(x, y)$ and their mirrored values $(1-x, 1-y)$ are encoded into spike times, such that a larger value of the coordinate results in a later spike time, and an early spike time for its mirrored version.

\linelabel{rev:refArchitecture}
The network architecture as shown in~\cref{fig:intuition} is a feed-forward multi-layer configuration with four input neurons, followed by a variable-size neuron layer (hidden layer) and finally an output layer, comprising three neurons for the three classes 
\linelabel{rev:deepNetsRef}
(a study on deeper networks is provided in~\cref{sec:si_deeper_networks}).
Delay layers are inserted between neuron layers, as previously illustrated in the computational graph (\cref{fig:computationalGraph}).
The neurons have no configurable biases, and the time constants are configured such that $\taum=2\taus$.
Thus, we utilize~\cref{eq:doubleTimeEquation} for training.
The refractory period $\tauref$ is set to infinity, such that all neurons only spike once. 
The output is represented in a \gls{ttfs} decoding scheme, where the first output neuron to spike indicates the predicted class for a given input.
To avoid negative or excessively large values for the delays, the effective delay $d$ is calculated as a logistic function of a trainable parameter $\theta_d$ such that $d = \lambda \cdot \sigma(\theta_d)$, which ensures that the delays
remain bounded between $0$ and $\lambda$. Further details can be found in \cref{sec:si_extended_simu}.

We have chosen the time-invariant \gls{mse} loss to improve accuracy and stability of training: 
\begin{equation}
    \mathcal{L}_{\Delta\text{MSE}}[\vect{t}, n^\star; \Delta_t] = \frac{1}{2} \sum_{n \neq n^\star} \left[ (t_n - t_{n^\star}) - \Delta_t \right]^2 \; ,
\end{equation}
where $n^\star$ and $n$ denote the respective indices of the correct and wrong label neurons and $\Delta_t$ is a freely chosen parameter. 
The purpose of introducing $\Delta_t$ into the loss is to achieve a specific separation of $\Delta_t$ between the spike times of the correct and incorrect label neurons, instead of providing precise target spike times.
To ensure a balance between model accuracy and hardware compatibility, $\Delta_t$ is set to $0.2\taus$ in our simulations.

\paragraph{Results}
We investigate the effects of different types of delay layers on accuracy, compared to configurations without any delays. \Cref{fig:simulation} reports the performance of our approach on the \gls{yy} dataset across different network sizes.
\linelabel{rev:testErrorDef}
\Cref{fig:simulation}b shows the percentage of misclassified samples in the test set (test error) of the network as a function of the number of hidden layers.
It demonstrates that co-training delays alongside the weights always improves performance, regardless of the specific type of delay.
Among the delay-augmented configurations, the
variant with synaptic delays outperforms the ones with axonal- or dendritic-only parameters.
This is in line with expectations, as synaptic delays offer the greatest configurable parameter space among the three delay variants.

\Cref{fig:simulation}c displays the same test errors, but now as a function of the number of parameters.
This representation reveals that, at least for the \gls{yy} dataset, delay-augmented networks with the same number of parameters perform similarly well, regardless of the type of delay.
As before, for a given number of parameters, the co-training of delays always yields at least as good results as the training of synaptic weights alone.
In other words, for the same memory footprint, a mix of both weights and delays is better than just synaptic weights.

Notably, the functional benefit of trainable delays depends to a great extent on the temporal structure of the data.
In particular, we expect the training of delays to have a larger impact if the input data spans longer time scales.
For \gls{yy}, it is straightforward to change the temporal volume occupied by the dataset by modifying its span -- the time difference between the earliest and latest possible input spikes.
\Cref{fig:simulation}d shows the effect of trainable delays across these different spans.
For small spans, errors are high because the temporal dynamics in the data are too fast for the intrinsic dynamics of the \gls{lif} neurons.
However, beyond a certain point, we always observe a clear benefit of co-training delays and weights as opposed to weights alone.
Furthermore, for a larger dataset span where input spikes can consequently be further apart, the range $\lambda$ of available delays that are able to push the \glspl{psp} together becomes increasingly relevant.

Optimal learning rates are determined through hyperparameter optimization for each configuration of neuron and delay layers.
Across all investigated settings, our approach demonstrates robust training convergence (\cref{fig:SI_extended_results}a) as well as exploitation of all available resources (\cref{fig:SI_extended_results}b).
Overall, these results clearly evince the added value of learning delays, as well as the ability of our algorithm to capitalize on this potential.

\section{Hardware results}\label{sec:hw_results}

\begin{figure*}[!t]
    \centering
    \includegraphics{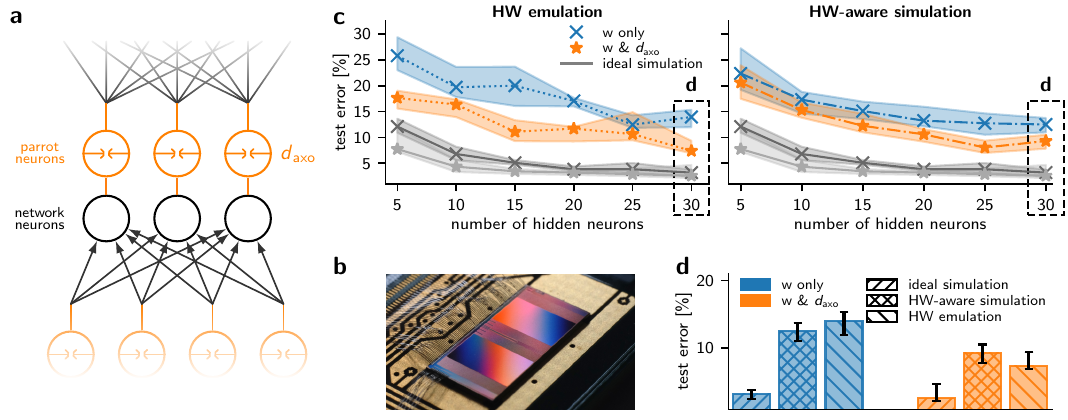}
    \caption{%
        \label{fig:hw_analog_parrot}
        \textbf{In-the-loop training with on-chip axonal delays on BrainScaleS-2.}
        \textbf{a)} Schematic illustration of the network architecture for on-chip axonal delays; here, we apply this generic approach to the BrainScaleS-2 neuromorphic hardware.
        Each neuron in the network (black) is paired with a parrot neuron (orange) connected in a one-to-one scheme.
        The parrot neuron repeats each of its input spikes with a configurable delay.
        \textbf{b)} Photograph of the BrainScaleS-2 neuromorphic chip (taken from~\cite{mueller2020bss2ll}).
        \textbf{c)} Median test errors and \gls{iqr} on the Yin-Yang dataset when training network weights and axonal delays (orange) or only weights (blue).
        The dash-dotted lines indicate a hardware-aware simulation (cf.\ \cref{sec:si_hardware_aware_sim}) and the dotted lines the hardware emulation results.
        For comparison, we also show the ideal software simulation results from \cref{fig:simulation}b in gray.
        The shaded areas indicate the \gls{iqr} over 10 runs with different seeds.
        The values for networks with \num{30} hidden neurons (highlighted by the dashed box) are shown for a better comparison in panel d.
        \textbf{d)} Detailed comparison of performances at 30 hidden neurons of an ideal simulation, hardware-aware simulation and emulation on neuromorphic hardware.
    }
\end{figure*}

To calculate the gradients for training weights and delays in \glspl{snn}, DelGrad only requires spike time recordings, compared to surrogate-gradient-based approaches, which also require recording membrane potential.
Therefore, DelGrad is ideally suited for implementation on a variety of neuromorphic substrates, whose output is spike-based, by design~\cite{Boahen1998}. 
Here, we demonstrate the flexibility of our method by describing a successful application in silico, on the neuromorphic platform \gls{bss2} that does not natively support delays.

The \gls{bss2} system (\cref{fig:hw_analog_parrot}b, \cite{billaudelle2020versatile, pehle2022brainscales2}) is built around a mixed-signal neuromorphic chip with $512$ physical neuron circuits.
The neuron dynamics are accelerated compared to biological time scales by a factor of $10^3$.
The neuron circuits emulate the dynamics of the \gls{adex} model~\cite{brette2005Adex} with individually configurable parameters for each neuron \cite{billaudelle2022accurate}.
Neighboring neuron circuits can be connected to form multi-compartment neurons~\cite{schemmel2017accelerated}.
The connectivity between the neurons on the chip can be configured arbitrarily within the constraints of the two $256 \times 256$ synaptic crossbar arrays. 
The synaptic weights are configured digitally with \SI{6}{\bit} resolution.

Although the current generation of \gls{bss2} does not natively support on-chip delays, we present an approach that allows us to explore the computational potential of delays for the current substrate.
We realize on-chip axonal delays by re-purposing a subset of the available neurons as delay elements.
For this, we utilize the adaptation circuitry as well as multi-compartment functionality of the neurons on chip.
This allows us to perform in-the-loop training of both synaptic weights and the on-chip axonal delays and illustrate the computational advantage obtained by the inclusion of delays.
The details of the delay implementation are provided in \cref{sec:si_adex_delays}.
We additionally provide a proof-of-concept of a different on-chip realization of axonal delays using \gls{lif} neuron dynamics, which is the most widely adopted neuron model for hardware platforms (see \cref{sec:si_lif_delays}).

\paragraph{Setup}

Even without an explicit hardware implementation of delays, an effective axonal delay can be achieved by exploiting the dynamics of the on-chip infrastructure.
For that, a ``parrot neuron'' is connected to the output of a neuron that is part of the actual trained network (\cref{fig:hw_analog_parrot}a).
For any spike that the network neuron produces, the parrot neuron is configured to elicit a spike after the desired delay.

In our implementation on \gls{bss2}, this behavior is achieved via the interplay between the two neuron compartments that form a parrot neuron.
The first compartment reacts to each incoming spike with a reset, which clamps the membrane voltage to the reset potential for a refractory period, during which the neuron is not responsive to incoming spikes.
After the end of the configurable refractory period, the second compartment becomes active and its adaptation mechanism almost instantly triggers an output spike of the parrot neuron.
Therefore, a spike is generated after the configurable refractory period, used here as the delay.
For a more detailed description of the mechanism see \cref{sec:si_adex_delays}.
We use this method, as it allows us to control the delay produced by the parrot neuron via its refractory period, which is digitally controlled on \gls{bss2} using 8 bits of precision.
This results in a more precise and easily configurable delay, compared to using an analog variable, and is likely closer to a future implementation of native delays on a \gls{bss2}-like system.

This delay mechanism allows us to train a network with axonal delays on \gls{bss2}.
We use an in-the-loop training approach, which means that we present a batch of inputs to the network on chip and record the spike times.
The spike times are sent back to the host computer where the loss and the backward pass are calculated in software.
The resulting updates for weights and delays are then used to reconfigure the chip before the next batch is presented.

\paragraph{Results}

With this chip-in-the-loop setup, we train and evaluate networks, with synaptic weights alone, as well as networks that incorporate both adjustable weights and axonal delays (\cref{fig:hw_analog_parrot}c).
Similar to the simulation results presented in \cref{fig:simulation}, we experimentally confirm an accuracy gain, over a range of network sizes, for the networks with additional delay parameters, compared to the ones with only weight parameters.

Overall, the final test errors reached in the software simulation are lower than the ones measured on hardware.
This is expected, as hardware effects such as trial-to-trial variations, fixed-pattern noise and jitter on the on-chip delays disturb the dynamics.
To illustrate and characterize these effects, we measured the magnitude of several noise sources found on the hardware and modeled them in a series of hardware-aware simulations.
\Cref{fig:hw_analog_parrot}c shows that, when the various sources of noise are realistically modeled based on hardware measurements, the hardware-aware simulations capture the increase in test error similarly to the actual emulation.
This confirms that the gap in accuracy between software simulations and hardware experiments is mostly due to the modeled sources of noise.
For an in-depth description of the noise models employed in the hardware-aware simulations and an analysis of the impact of the different noise sources on the network performance, we refer to \cref{sec:si_hardware_aware_sim}.

    For an easier comparison we focus on the most expressive networks with 30 hidden neurons, highlighted by boxes in \cref{fig:hw_analog_parrot}c, and collect the achieved test errors in \cref{fig:hw_analog_parrot}d which amount to \SI{7.40}{\percent} with axonal delays and \SI{13.95}{\percent} in the weight only case on the hardware.
    A full report of the achieved test errors and \gls{iqr}, both in hardware-aware simulation and on chip, can be found in \cref{tab:hw_aware_sim}.
Additionally, we note that the performance gap between the delay and no-delay setup is significantly wider on hardware than in the ideal software simulations.
We hypothesize that this effect arises because the \gls{yy} classification problem, by design, does not require a large network to solve, leading to few learnable parameters, low redundancy, and consequently a greater sensitivity to noise.
Introducing axonal delays increases redundancy, due to the higher parameter count.
However, this increase is rather small compared to the number of parameters in a weight-only setup. 
This suggests that the computational properties of the delays are at least partially responsible for making the network more noise resilient, explaining the larger performance gap between the two networks on noisy hardware.

These results illustrate that our method for training delays is not only applicable in ideal software simulations but can also be applied to mixed-signal neuromorphic systems. Additionally, they demonstrate the benefit of learnable delays for neuromorphic platforms, especially in resource-constrained scenarios, and might encourage the inclusion of delay mechanisms in future generations of neuromorphic systems.

\section{Discussion}
We have introduced an exact event-based algorithm for training temporal variables, specifically transmission delays, in conjunction with synaptic weights in \glspl{snn}.
Additionally, we have experimentally validated its effectiveness through both software simulations and neuromorphic hardware implementations.

Delay parameters were previously demonstrated to increase the representational power of the \glspl{snn}, even without optimization, just by training the weight parameters to select the useful delays for spatio-temporal feature detection~\cite{dagostino_etal2024_denram,patino_etal2023_imec_delay,habashy2024diverseNeurons}.
However, this optimization-through-selection approach requires an over-allocation of resources in order to provide a sufficiently diverse set of delay parameters from which the best can be selected.
To illustrate this, we compare a network of random fixed delays to a network with trained delays using DelGrad. 
We show in \cref{fig:SI_extended_results_randButFixed} that for the same number of delay parameters, the network with trained delays and weights has a clear accuracy advantage over the network with randomly initialized delays with weight-only optimization.
Therefore, it is advantageous to combine a dedicated learning algorithm for transmission delays with hardware capable of configuring them accordingly.

Algorithms based on surrogate gradients for direct training of delay elements have been explored recently, using temporal convolution kernels~\cite{hammouamri2023learning} or numerical solutions that estimate the delay gradients using finite-difference approximations~\cite{shrestha2018SLAYER}.
\linelabel{rev:refDelayFiniteDiff}
However, as pointed out in~\cite{hammouamri2023learning}, delay training based on finite-difference approximation~\cite{shrestha2018SLAYER} appears to not be sufficiently accurate to achieve an improvement over fixed, random delays.
Additionally, both approaches use a time-stepped framework for calculating the gradients.

As such, delays are represented implicitly in the number of simulation time steps before transmitting a spike.
However, as delay parameters are essentially shifts in individual spike times,
we argue that it is more natural to have a framework where the information is explicitly represented by these spike times~\cite{madhavan2021temporal,goeltz2021fast,wunderlich2021eventprop,Stanojevic2024GerstnerBellec,Schrauwen2004ExtendingSpikeProp}, and delays are learned as additive parameters for these times.
Furthermore, the objective of building efficient asynchronous neuromorphic systems, where time represents itself, is an additional motivation for representing temporal information in spike times~\cite{mead1990neuromorphic}.
Such representations are naturally available from event-based sensors, where the change in the signal is encoded into spike times using the delta modulation encoding scheme~\cite{gallego_etal2020_dvs,vanschaik_etal_cochlea,bartolozzi2021neuromorphic,}.

This work brings together all the aforementioned objectives: DelGrad presents an event-based framework for gradient-based co-training of delay parameters and weights, without any approximations, and which meets the typical demands and constraints of neuromorphic hardware, as demonstrated experimentally on an analog mixed-signal neuromorphic system.
As such, it takes an important step towards fully exploiting the temporal nature of \glspl{snn} for memory- and power-efficient end-to-end event-based neuromorphic systems.

\paragraph*{Different delay types and hardware considerations}

In this work, we have also compared the effect of dendritic, axonal and synaptic delays on the performance of \glspl{snn} on a representative task.
The synaptic delays have the highest impact on increasing the expressivity of \glspl{snn}, compared to using only dendritic or axonal delays. 
However, from a hardware perspective, the addition of synaptic delays imposes a quadratic growth on the size and thus the on-chip area of the network, compared to a linear growth in the case of axonal and dendritic delays.
In fact, we find that when comparing the performance for equal parameter counts, the gap between different types of delays vanishes while the superiority over weight-only training persists.
As memory represents one of the most critical constraints on hardware, reducing on-chip memory is of utmost importance.
In particular, this means that a redesign of a chip with fewer neurons but with an intrinsic delay mechanism, based on our findings, will save energy while maintaining expressivity and performance.
Therefore, our work suggests that it might be practical to consider
using dendritic or axonal delays in future hardware designs, combining favorable scaling and improved processing power.

\paragraph*{Hardware mapping}
\linelabel{rev:refHardwareMapping}
DelGrad provides an advantage in terms of hardware mappability as it only requires recording the spike times from on-chip neurons. 
This is in contrast to other approaches~\cite{hammouamri2023learning, shrestha2018SLAYER}, which need access to the membrane potential of all neurons  for surrogate gradient learning~\cite{cramer2022surrogate,goeltz2023gradients}.
Such voltage-based plasticity requires additional components for voltage readout, communication and potentially analog-to-digital conversion.
Furthermore, this information is much denser in time than the spikes themselves, imposing further stress on the overall communication bandwidth.
In both chip-in-the-loop and on-chip training scenarios, these additional requirements ultimately translate to additional circuitry; not only does this increase the complexity of the chip design, but it also inherently reduces the maximum implementable network size for a chip of a given area.
Moreover, if learning is to be implemented on-chip, this additional circuitry will negatively affect the device's energy efficiency during training.

\paragraph*{Outlook}
\linelabel{rev:refOutlook}
In this work, the \gls{yy} dataset was used as a proof of concept and first step to benchmark our approach.
The \gls{yy} dataset provides a problem that can not be solved linearly and where the information can be presented using a \gls{ttfs} encoding, similar to the previous work~\cite{goeltz2021fast}.
As indicated by our experiments in \cref{fig:simulation}d, the performance boost provided by the inclusion of learnable delays increases when the temporal features of the data span larger time scales.

Therefore, the natural next step will reside in a more thorough benchmarking on larger datasets, and in particular on data with explicit temporal components, such as~\cite{Cramer2022SHD, warden2018speech}.
Especially for data provided by event-based sensors, longer time scales are required and \gls{ttfs} might reach its limits as a feasible coding scheme.
Although our current software implementation only takes into account a single spike per neuron during the training, this is not a limitation of our proposed mathematical framework and training scheme (see~\cref{sec:si_math_multi}).
Additionally, the extension to more complex spike timing codes can go hand in hand with a shift from a feed-forward to a recurrent network architecture.

\FloatBarrier
\newpage
\AtNextBibliography{\small}
\printbibliography

\newpage
\section*{Acknowledgments and Funding}
We want to thank the EIS-Lab, NeuroTMA, CompNeuro, and ElectronicVision(s) groups, in particular Yannik Stradmann, Robin Heinemann, Joscha Ilmberger and Eric Müller, for the continuing support.
Additionally, we are grateful to the NNPC conference 2023 where this fruitful collaboration was initialized, as well as the CapoCaccia workshop where this work was first presented and received much helpful feedback especially from Paolo Gibertini and Maryada.
We also thank Guillaume Bellec for critical comments on the preprint and Florent Draye for providing feedback on the mathematical proofs.

The presented work has received funding from
the Manfred Stärk Foundation,
the EC Horizon 2020 Framework Programme under grant agreement 945539 (HBP) and Horizon Europe grant agreement 101147319 (EBRAINS 2.0),
the Deutsche Forschungsgemeinschaft (DFG, German Research Foundation) under Germany’s Excellence Strategy EXC 2181/1-390900948 (Heidelberg STRUCTURES Excellence Cluster),
Swiss National Science Foundation Starting Grant Project UNITE (TMSGI2-211461),
and the VolkswagenStiftung  under grant number 9C840.
\clearpage
\appendix
\onecolumn
\AddToHook{cmd/section/before}{\FloatBarrier}
\AddToHook{cmd/subsection/before}{\FloatBarrier}
\AddToHook{cmd/subsubsection/before}{\FloatBarrier}
\glsresetall
\renewcommand{\thesubsection}{SI.\Alph{subsection}}
\renewcommand{\thefigure}{SI.\arabic{figure}}
\setcounter{figure}{0}
\renewcommand{\thetable}{SI.\arabic{table}}
\setcounter{table}{0}

\section*{Supplementary Information}
\addcontentsline{toc}{section}{Supplementary Information}
\addtocontents{toc}{\protect\setcounter{tocdepth}{0}}

\subsection{Additional simulation results}\label{sec:si_extended_simu}

\subsubsection{Deeper networks}\label{sec:si_deeper_networks}

\begin{figure}[ht]
    \centering
    \includegraphics{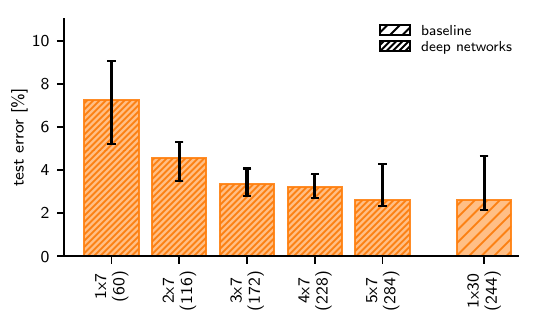}
    \caption{%
            \label{fig:deeper_networks}
            \textbf{Test error for networks of varying depth.}
            Networks of the form $l\times n$, with $l$ the number of hidden layers of $n$ neurons.
            The number of parameters is indicated in parentheses.
            The test error decreases from a depth $l=1$ to $l=5$ and constant width of $n=7$.
            The 1$\times$30 baseline lies between the 4$\times$7 and 5$\times$7 configurations.
    }
\end{figure}

DelGrad maintains its performance when scaling to deeper networks.
\Cref{fig:deeper_networks} shows a series of networks including axonal delays with increasing depth (from $1$ to $5$ hidden layers) and fixed width ($7$ neurons per hidden layer), demonstrating a consistent decrease in test error.
The baseline configuration with $1$ hidden layer of $30$ neurons falls between the 4-layer and 5-layer networks, both in terms of test error and number of parameters.

\subsubsection{Ablation studies}\label{sec:si_ablation_studies}
\begin{figure}[ht]
    \centering
    \includegraphics{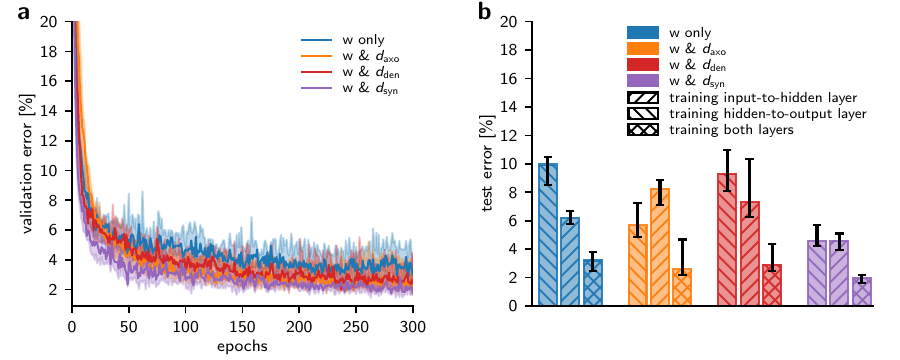}
    \caption{%
        \label{fig:SI_extended_results}
        \textbf{Extended simulation results.}
        \textbf{a)} Comparison of the validation error during training for the different delay types and a network without delays.
        All networks have one hidden layer with $30$ neurons.
        \textbf{b)} Ablation study (for networks with a hidden layer size of 30 neurons) on the effect of only training the connections between input and hidden layer or between hidden and output layer.
        }
\end{figure}

\begin{figure}[t]
    \centering
    \includegraphics{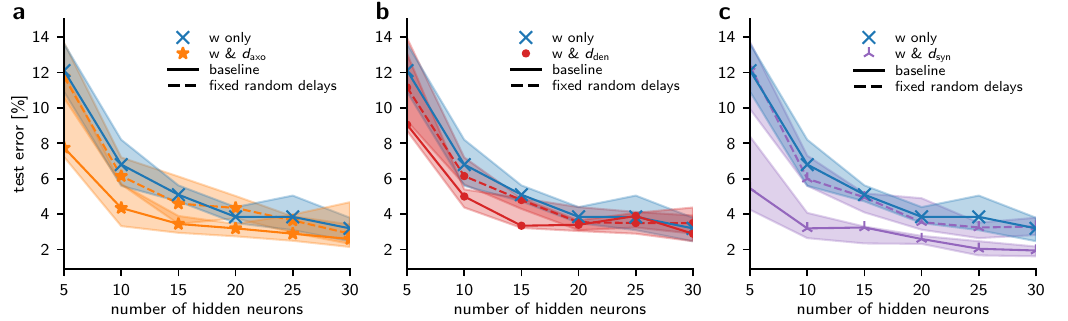}
    \caption{%
    \label{fig:SI_extended_results_randButFixed}
        \textbf{Comparison to random but fixed delays.}
        For a setup of axonal (a), dendritic (b) or synaptic (c) delays the performance of fully trained networks (i.e.\ learning weights and delays) are compared to networks with random but fixed delays (dashed lines) and networks without delays (blue). 
    }
\end{figure}

In addition to the results presented in \cref{sec:sim_results} we performed several ablation studies that illustrate the effect of trainable delays in our networks.
\Cref{fig:SI_extended_results}b demonstrates that the training makes use of all available resources to improve the task performance by comparing fully trained networks with networks where weight and delay training is disabled either for the input-to-hidden layer connections or for the hidden-to-output ones.
The fixed parameters are initialized by sampling from a Gaussian distribution that approximates the empirical distribution of weights and delays observed in a fully trained and optimized baseline network.
This ensures that the parameters lie in an appropriate range to solve the task.
On the one hand, this shows that the training of all layers is required to obtain optimal performance on the task. On the other hand, it also proves that in the full training, useful gradients are provided to all parameters in all layers.

Finally, we compare the performance achieved when we train weights and delays to an approach similar to the one employed in \cite{dagostino_etal2024_denram}, where weights are trained, but the delays are fixed and random (\cref{fig:SI_extended_results_randButFixed}).
The mean and standard deviation for configuring the delays are obtained from a hyperparameter search on a wide range of values.
We see that for this task, training the delays compared to just providing a random selection that is not adjustable provides a clear advantage for all delay types and especially in smaller networks.

\subsubsection{Hardware-aware software simulations}\label{sec:si_hardware_aware_sim}

\begin{figure}[!t]
    \centering
    \includegraphics{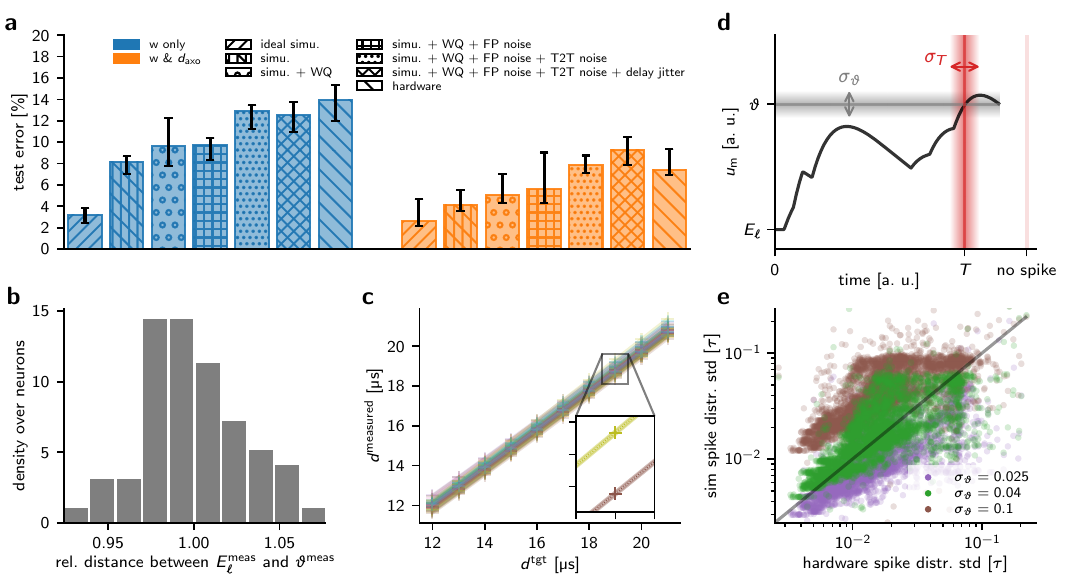}
    \caption{%
        \label{fig:si_hw_aware_sim}
        \textbf{Hardware-aware simulation and noise measurements}
        \textbf{a)} Ablation study to disentangle the impact of different hardware effects on training performance.
        In addition to the ideal software simulation (leftmost bar) and the actual hardware result (rightmost bar), it shows hardware-aware simulation results with progressively more hardware effects (from left to right): weight quantization (WQ), fixed-pattern noise (FP), trial-to-trial noise (T2T) and delay jitter.
        In principle, for each setting slightly different hyperparameters would be optimal; however, for a more direct comparison within reasonable simulation time, we've selected the set of hyperparameters that have proven reliable in the actual hardware emulation for all experiments starting from the second bar (simu.).
        Bar height indicates median test error and the error bars show the \gls{iqr}.
        The values shown correspond to \cref{tab:hw_aware_sim}.
        \textbf{b)} Histogram of the normalized relative distance between the measured threshold $\Vth$ and the leak potential $\Vleak$.
        This distribution is used to estimate the fixed-pattern noise between different neuron circuits on the hardware.
        \textbf{c)} Measurements of delays produced by parrot neurons on \gls{bss2} over the corresponding target values used to configure the parrots.
        \textbf{d)} Sketch of the non-linear relationship between variations in the threshold (gray area) and the resulting variations in the output spike timing (red area).
        \textbf{e)} Comparison of variations on the output spike timing on hardware and in the hardware-aware simulation for different assumed trial-to-trial noise magnitudes $\sigma_\vartheta$. Each point represents the variations on the spike time for one neuron that was presented multiple times with the same input pattern. The plot combines data for multiple neurons and different input patterns.
    }
\end{figure}

The training of \glspl{snn} on mixed-signal neuromorphic hardware brings several additional challenges compared to an ideal software simulation.
These challenges include, among others, limited resolution and ranges on parameters such as the synaptic weights, fixed-pattern noise and trial-to-trial variability.
The impact of these factors on the final outcome of the training difficult to predict and disentangle.
To nevertheless get an impression of this, we attempt to mimic these effects in a software simulation.
We call this hardware-aware training.
With the hardware-aware simulation we can perform an ablation-study that allows us to step-by-step include more levels of hardware-realism and observe the effect on performance.
We ensure that the noise levels and restrictions that we include in the simulation are of similar magnitude than what is encountered on a mixed-signal platform.
This of course strongly depends on the choice of neuromorphic system and we show it here for our configuration of \gls{bss2}.

\paragraph{Weight ranges and quantization}

The synaptic weights on \gls{bss2} have a \SI{6}{\bit}-resolution.
To model this in software we use the same approach as described in detail in the methods of \cite{goeltz2021fast}:
We match the available maximal \gls{psp} height in simulation and on hardware and then train with quantization-aware training within the available range.

Due to the limited weight range, the maximum achievable weight is constrained, requiring, for our chosen neuron parametrization, multiple input spikes to reliably trigger an output spike in a neuron.
To address this, as described in~\cite{goeltz2021fast}, we replicate each input spike across five channels.
This approach effectively increases the maximum achievable weight by a factor of five, a technique we refer to as ``channel multiplexing''.
While these multiplexed channels introduce additional parameters to the network, in the weight-only training they do not enhance its computational capacity.
This is because all multiplexed channels share the same input and hence, for each input spike the actual input into the hidden layer is simply the sum of the weights across all channels.
In contrast, when delays are available, each channel can have a unique delay, resulting in input spikes arriving at different times. These staggered spike timings separate the influence of each channel, allowing the individual weights to have a distinct effect and thereby increasing the computational capacity of the delay-based network compared to the weight-only ones.
However, since this computational advantage arises solely from hardware limitations, we enforce shared delays across multiplexed channels in the delay-based network to ensure a fair comparison.

For the small networks with a hidden layer size of only 5 neurons we encounter the same problem of having too few spiking neurons in the hidden layer to reliably activate the output layer.
On \gls{bss2} we can solve this by using two synapse circuits instead of one for a connection, effectively doubling the weight (the delays here are shared automatically, because the spikes are received from a parrot neuron).
This requires more chip resources per connection, which is why typically this is avoided, but for the smallest networks it proved to be necessary.
We mimic this in the hardware-aware simulation, for the hidden layer size of 5, by increasing the available weight range by a factor of 2.

\paragraph{Fixed-pattern noise}

Fixed-pattern noise, or sometimes called frozen noise, is caused by imperfections in the chip fabrication process and causes slight differences between the neuron circuits.
This results in the dynamics of each neuron on chip to differ slightly.
These differences between neurons are static over time.
The effects of fixed-pattern noise can partially be mitigated using calibration procedures, which we employ, but some variability between the neurons remains.
Neuron parameters that are affected by fixed-pattern noise are for example the time constants, the strength of the synaptic input, the resting potential and the threshold.
As a simplification for our simulations we do not model all sources of fixed-pattern noise individually, but summarize them all into one source.
The variation between neurons of the difference between resting potential and threshold is easy to measure on chip (\cref{fig:si_hw_aware_sim}b).
Therefore, in our simulations, we model the fixed-pattern noise based on this parameter. 
In the hardware-aware simulations, at network initialization, a random offset to the threshold of each neuron is drawn from a Gaussian distribution and applied for the whole training procedure.
For an estimate on the variance of the Gaussian, we use the variance observed in \cref{fig:si_hw_aware_sim}b and increase it slightly to account for the other sources of fixed-pattern noise.

\paragraph{Trial-to-trial variability}

In addition to fixed-pattern noise, which is static over time, we also observe trial-to-trial variability on the hardware.
Electronic circuits are affected by temporal noise on all timescales.
High frequency components become apparent as visible jitter on top of the underlying signal, for example on membrane traces.
Lower frequency components, in contrast, occur also on timescales often greater than individual observation periods and can thus manifest themselves as pseudo-static offsets of a signal on a trial-to-trial basis.
These low frequency components are particularly strong due to the fact that the noise characteristics of electronic devices are typically dominated by flicker noise with a spectral density, or ``amplitude'', proportional to $ 1/f $.

The resulting trial-to-trial variability can be interpreted as random fluctuations of neuron parameters on comparatively long time scales:
We model it by varying the neuron parameters between experiments (i.e.\ different batches) but assume them to stay constant within experiments.
Most noise sources -- such as the leak and threshold terms but also the synaptic integrators and input circuits -- eventually affect the neuronal membrane state in form of a random offset and we thus subsume all of them in a variation of the distance between leak and threshold potentials.
Therefore, if the same neuron on the same chip is presented with the same input in different batches, its output spike time will differ slightly.
The relationship between a variation on the threshold and the resulting variations on the spike times of the neuron is highly non-linear (\cref{fig:si_hw_aware_sim}d).
This makes the trial-to-trial variability hard to model directly on the spike times, even though this is where it is observed.
Instead, we choose to add random offsets to each threshold of every neuron for every batch, which automatically then approximates the trial-to-trial noise observed on the spike times on the hardware.
To confirm this, we repeatedly present the same batches of samples to the hardware and record all occurring spikes.
Then we present the same samples repeatedly to the hardware-aware simulation, where for each batch and neuron, we add a random sample as the offset to the threshold, drawn from a Gaussian centered around zero and with width $\sigma$. 
For the right choice of $\sigma$ we observe that the variances, observed for each sample over many repetitions, match well between hardware-aware simulation and experiments (\cref{fig:si_hw_aware_sim}e).

\paragraph{Delay effects}

\Cref{fig:si_hw_aware_sim}c provides an estimate of how accurately our parrot neuron setup reproduces a target delay.
While the variations across multiple trials are minimal, we account for them in the hardware-aware simulations.
Specifically, we model the variability of the delay circuits by introducing Gaussian noise ($\sigma=\SI{0.01}{\tau}$) to the output spike times of the delay layers. 
In general, spike signal communication on \gls{bss2} is highly reliable, and spike loss only becomes a concern at very high firing rates within the network.
However, the inclusion of parrot neurons for delay implementation introduces a potential new source of spike loss.
We measured the average rate of spike loss caused by the parrot neuron setup and found it to be negligibly small. 
Consequently, this effect is not incorporated into our simulations.

\paragraph{Results}

With the hardware-aware simulation setup described above we perform an ablation study and compare the results to the actual hardware emulation as well as the ideal software simulation (\cref{fig:si_hw_aware_sim}a and \cref{tab:hw_aware_sim}).

Some increase in error is introduced when transitioning from the hyperparameters optimized for the ideal simulations to those used on our hardware. 
This adjustment was necessary because the optimal parameters identified by software-based \gls{hpo} did not perform well on the hardware.
In software simulations, hyperparameters were optimized independently for each network size and delay configuration (e.g., axonal or synaptic). However, such extensive \gls{hpo} is impractical on hardware due to the inability to parallelize runs on the chip, making the process prohibitively time-consuming.
To address this, we optimized a single set of hyperparameters that performs reasonably well across both weight-only and axonal delay scenarios and for all network sizes.
While this compromise set does not achieve the same performance as the highly optimized software case, it is a good trade-off between the feasibility and performance.
More importantly, as shown in \cref{fig:si_hw_aware_sim}a and \cref{tab:hw_aware_sim}, all modeled noise sources contribute to the increased error, and the final error in hardware-aware training matches closely with the results observed on the chip. This demonstrates that, despite significant simplifications in modeling hardware effects, our measurement-based estimation of noise sources effectively captures the chip’s general behavior.

\input{table_hw_aware.tex}

\subsubsection{Comparison to literature}\label{sec:si_literature}

\begin{figure}[ht]
    \centering
    \includegraphics{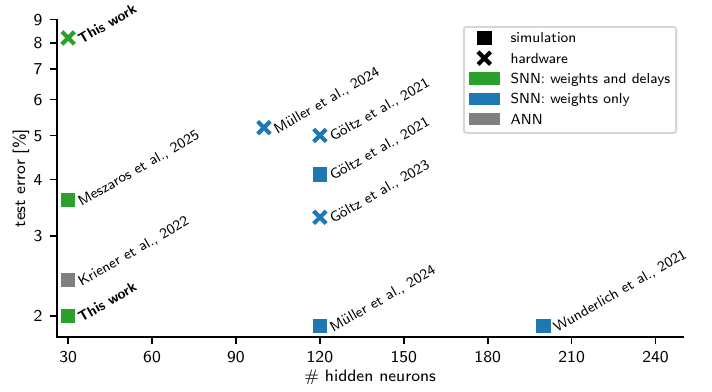}
    \caption{%
        \label{fig:SI_yy_comparison}
        \textbf{Comparison to other results on the \gls{yy} dataset.}
        Each data point represents the mean test accuracy achieved on the \gls{yy} dataset for a certain size of the hidden layer.
        Simulation results are marked as squares while hardware results are plotted as crosses.
        Results employing delays are colored in green, while weight-only networks are blue.
        The data is collected from the following publications: Meszaros et al., 2025~\cite{meszaros2025efficient}; Kriener et al., 2022~\cite{kriener2021yin}; Müller et al., 2024~\cite{muller2024jaxsnn}; Göltz et al., 2021~\cite{goeltz2021fast}; Göltz et al., 2023~\cite{goeltz2023gradients} and Wunderlich et al., 2021~\cite{wunderlich2021eventprop}.
    }
\end{figure}

\subsection{Hardware implementation and additional results}\label{sec:si_axonal_details}

As \gls{bss2} does not include dedicated circuitry for emulating delays, we re-purpose neuron circuits to act as delay elements (parrot neurons, see \cref{fig:hw_analog_parrot}a).
For any incoming spike the parrot neuron is configured to produce an output spike with a certain controllable delay, as described below.

\subsubsection [Axonal delays using AdEx and multi-compartment functionality]%
    {Axonal delays using AdEx and multi-compartment functionality}%
    \label{sec:si_adex_delays}

\begin{figure}[!t]
    \centering
    \includegraphics{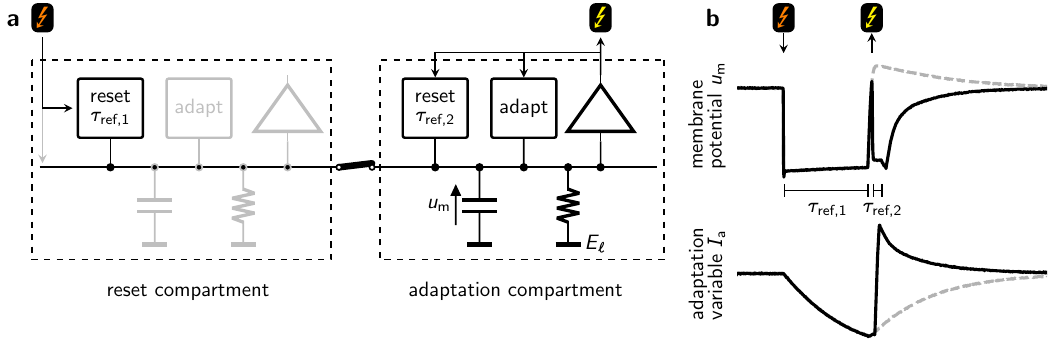}
    \caption{%
        \label{fig:hw_adex_delays}
        \textbf{Schematic illustration of the parrot neuron used to implement delays on BrainScaleS-2.}
        \textbf{a)} Sketch of the two neuron compartments (dashed boxes) used for the delay mechanism.
        The reset and adaptation circuits are drawn as rectangles, the threshold comparator as triangles.
        Disabled components are drawn in gray.
        Lightning bolts indicate incoming (orange) and outgoing (yellow) spikes.
        \textbf{b)} Recorded traces of membrane voltage and adaptation when the parrot neuron receives an input spike and produces a delayed output spike.
        The dashed gray traces show the circuit with disabled threshold comparator in the second compartment.
    }
\end{figure}

To obtain the results shown in the main text the delays are implemented using the multi-compartment and the adaptation functionality~\cite{brette2005Adex} of the \gls{bss2} hardware~\cite{pehle2022brainscales2, billaudelle2022accurate}.
\Cref{fig:hw_adex_delays} illustrates the mechanism that consists of two separate neuron compartments, from here on called reset and adaptation compartment, corresponding to their respective role.
The membrane capacitance of the reset compartment is disabled and the two compartments are short-circuited such that they share a membrane voltage.
The dynamics of the adaptation compartment are described by the equations of the \gls{adex} model with disabled exponential component:
\begin{align}
    \taum \umdot(t) &= [\Vleak - \um(t)] + \Is(t) / \gl - I_\text{a}(t) / \gl \\
    \tau_\text{a} \dot{I}_\text{a}(t) &= a (u - \Vleak) - I_\text{a}(t)
\end{align}
where the adaptation variable $I_\text{a}$ is a leaky integrator that is driven by the difference between the membrane voltage and the leak potential.
Note that no synaptic input is connected to this compartment, and thus $\Is(t)$ is zero at all times.
When $\um(t)$ crosses the spiking threshold at $t_\text{spike}$, the membrane voltage is reset and clamped to $\Vreset$ for the duration of the refractory period $\taureftwo$ and spike-triggered adaptation causes a jump on the adaptation variable
\begin{align}
    u(t) &= \Vreset \;\; \forall t \in (t_\text{spike}, t_\text{spike} + \taureftwo] \\
    I_\text{a} &\rightarrow I_\text{a} + b
\end{align}
where $b$ is the parameter controlling the magnitude of the spike-triggered adaptation.

An input spike arriving at the reset compartment of the parrot neuron triggers an immediate reset which clamps the membrane potential (shared between both compartments) to the low reset potential $\Vreset$ for the duration of $\taurefone$.
Since $\Vreset$ is lower than $\Vleak$, the magnitude of the adaptation current $I_\text{a}$ in the adaptation compartment builds up during the refractory period.
Once the refractory period $\taurefone$ ends, the membrane voltage, now driven by the adaptation variable, can evolve freely away from $\Vreset$ and the accumulated strong adaptation current $I_\text{a}$ causes the membrane voltage to rapidly increase towards the threshold.
This phenomenon, although commonly caused not by a low $\Vreset$ but by inhibitory input, is known as inhibitory rebound:
The membrane potential overshoots the leak potential and becomes high enough to reach the spiking threshold of the adaptation compartment and produce an output spike, triggering the reset mechanism of this compartment.
Additionally, the spike triggers the spike-triggered adaptation mechanism of the \gls{adex} model, which causes a jump on $I_\text{a}$ by $b$, bringing back the adaptation $I_\text{a}$ close to zero.
Finally, the refractory period of the adaptation compartment $\taureftwo$ is configured to be very short and the membrane can almost instantly relax to the leak potential again.
After this, the parrot neuron is ready to receive and delay the next input spike.

As the inhibitory rebound causes an output spike immediately after the end of the refractory period of the reset compartment, the delay of this parrot neuron can be configured via $\taurefone$.
This is advantageous, as $\tauref$ is an \SI{8}{\bit} digital parameter on \gls{bss2} and can be configured to a large range of possible refractory periods without fixed pattern noise.
As shown in \cref{fig:si_hw_aware_sim}c, the delays can be reliably configured in the range of \SIrange{12}{21}{\micro\second}.
In principle, the available delay range is larger (approximately \SIrange{8}{35}{\micro\second}), but since we do not require such a large range of delays for our experiments and the values in the middle of the available range are the most stable, we restrict ourselves to what is shown in \cref{fig:si_hw_aware_sim}c.
Note that having the smallest available delay unequal to zero is not computationally relevant, as long as this minimal value is equal for all neurons.

\subsubsection{Proof-of-concept for axonal delays using only LIF dynamics}\label{sec:si_lif_delays}

\begin{figure}[!t]
    \centering
    \includegraphics{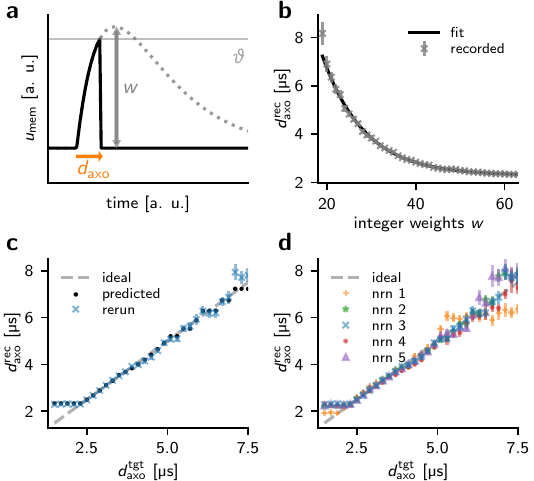}
    \caption{%
        \label{fig:hw_lif_delays}
        \textbf{Proof of concept for implementing on-chip axonal delays using only LIF dynamics on BrainScaleS-2.}
        \textbf{a)} Example membrane trace of a parrot neuron where the rise time of the \gls{psp} causes a delay of its output spike with respect to the output spike time of its afferent network neuron.
        This delay can be configured through an appropriate choice of the synaptic weight between network neuron and parrot neuron.
        \textbf{b)} Example BrainScaleS-2 recording of the relationship between the measured input-output delay of a parrot neuron $d_\text{rec}$ and its afferent synaptic weight (neuron 3 in d)).
        Mean and standard deviation are shown over $10$ runs with $50$ spike pairs each.
        An exponential fit (black) yields the calibration curve for the weight-delay relationship.
        \textbf{c)} Test of the calibration for the same parrot neuron as in b).
        The calibration curve from the fit in d) is used across a range of target delays $d_\text{axo}^\text{tgt}$ to determine the corresponding optimal synaptic weights.
        With this weight the delay is re-measured for 5 runs with 50 spike pairs each, checking the deviation between the predicted delay (black) and the actual recorded delay $d_\text{rec}$ (mean and standard deviation in blue).
        \textbf{d)} Same as c) but for 5 different parrot neurons on the chip to illustrate the variability between different neuron circuits.
    }
\end{figure}

Since \gls{lif} neuron models are more widely available on various neuromorphic platforms compared to the multi-compartment \gls{adex} model, we also present a proof-of-concept of how \gls{lif} models can be adapted for implementing analog on-chip delays, ensuring our results are more easily reproducible.

\paragraph{Setup}
The network setup in this method is similar to the previous section in that each network neuron has a corresponding parrot implementing the delay; However, the method of creating this delay is different:
We leverage the fact that due to the finite rise time of the \gls{psp} on the parrot's membrane voltage, its spike is delayed compared to the one of the network neuron (\cref{fig:hw_lif_delays}a).
The magnitude of this delay, which emulates the axonal delay of the network neuron, depends on several parameters, such as the synaptic weight $w$ of the connection between the network and parrot neurons, the time constants $\taus, \taum$ and the difference between threshold and leak potential of the parrot neuron.
For a theoretical description of the relationship between the parrot's delay and the synaptic weight see \cref{sec:si_lif_delay_theoretical}.

For our implementation of this scheme on \gls{bss2}, we control the delay solely via the synaptic weight $w$, keeping the time constants and potentials fixed:
while those parameters can be individually tuned, that (analog) configuration is slower compared to the (digital) weight setting.
Since in our trained networks on \gls{bss2} (\cite{goeltz2021fast} and \cref{fig:hw_analog_parrot}) we use neuron time constants of $\taus = \taum \approx \SI{6}{\micro\second}$, we aim to reach delays of the same magnitude here.
To achieve this, we configure the parrot neurons to have a synaptic time constant of $\taus = \SI{10}{\micro\second}$ and a membrane time constant of $\taum = \SI{15}{\micro\second}$.
A long refractory time of \SI{16}{\micro\second} ensures that each input to the parrot only triggers one output spike.

During the training of a network, it is required to reconfigure the parrot neurons on the chip to produce the correct axonal delays $d_\text{axo}^\text{tgt}$.
For this, the relationship between the synaptic weight $w$ and the produced delay $d_\text{axo}^\text{rec}$ needs to be measured.
Due to the usual variations in the manufacturing process (fixed-pattern noise), the on-chip analog neuron circuits are not exactly identical to each other.
Therefore, for every parrot neuron, we perform a separate calibration measurement in order to determine the precise mapping between the weight parameter and the recorded delay individually.
To this end, we configure a range of different weights and record the resulting delays $d^\text{rec}_\text{axo}$ (\cref{fig:hw_lif_delays}b for an example neuron). %
The full available weight range from $0$ to $63$ is not used, as for the lower weights, the parrot neuron does not reliably produce output spikes.
To include both temporal drift during one trial and trial-to-trial variations, we record delays during $10$ runs with $50$ pairs of input and output spikes and average the results.

We fit an exponential function $d(w) = \alpha + \beta \exp(\gamma (w + \delta))$ to the measured data.
The inverse of the fit function thus determines the relation between the target delay $d_\text{axo}^\text{tgt}$, and the optimal integer weight to configure on the chip.
To test the quality of the weight-delay fit, it is used to configure the chip for a whole range of target delays $d_\text{axo}^\text{tgt}$, while measuring the actual value of the delays produced on the chip $d_\text{axo}^\text{rec}$ (\cref{fig:hw_lif_delays}c).
The above process is repeated for 5 different neuron circuits on the chip, and  the results are compared to illustrate the impact of fixed-pattern noise in \cref{fig:hw_lif_delays}d.

\paragraph{Results}
\Cref{fig:hw_lif_delays}c shows the desired correspondence between the target, $d_\text{axo}^\text{tgt}$, and the recorded, $d_\text{axo}^\text{rec}$, on-chip delay values, especially in the intermediate delay range.
This underpins the feasibility of our proposed approach.
We note a plateau in the recorded delay values for lower targets, caused by the maximum possible on-chip weight value, corresponding to the shortest possible delay.
Additionally, a larger deviation and more instability is observed for larger target delays; this effect has two causes:
a worse quality of the exponential fit and an increased instability of the threshold crossing in the region where the \gls{psp} plateaus.
Such increasing instability is unavoidable in analog neurons, as any amount of noise on the membrane voltage results in increased trial-to-trial variability when the peak of the \gls{psp} is close to the threshold.

Although each neuron can be configured individually to produce the desired behavior, there is still some variability between the neurons.
However, we do not expect these variations to be harmful in practice; in fact, some heterogeneity between neurons behavior might even be beneficial, as has been shown in~\cite{Perez_Nieves_2021}.

While the presented idea can be feasible for small networks and tasks, it is clearly suboptimal, as it requires a portion of the available neuron circuits to be used as delay elements instead of their usual role in the network.
Additionally, several practical considerations have to be taken into account when this setup is included in the training of a full network.
First, the range of achievable delays is limited by the time constants of the parrot neurons.
In our experiments we targeted a delay range of approximately \SI{6}{\micro\second} which corresponds to the delay range used in the results in \cref{sec:sim_results},
Second, for a correct delay on an incoming spike, the parrot neuron's membrane voltage and synaptic currents need to be at their resting values.
Therefore, the interval between spikes arriving at the parrot neuron needs to be large enough, which can be ensured by increasing the refractory time of the neurons in the network.
However, for tasks where the neurons need short refractory periods to process their input correctly, this method of producing axonal delays is not suitable.
Nevertheless, these results provide a proof of concept that axonal delays can be implemented by repurposing resources and circuits that are universally available on most neuromorphic substrate.

\subsection{Complete equation for the time of the first spike}\label{sec:si_math_ab}
Given a sequence of input spikes $\{t_i\}$ and corresponding weights $\{w_i\}$, we define
\begin{align}\label{eq:SI_mathAB}
    a_n := \sum_{i \in C} w_i \exp\left(\frac{t_i}{n\taus}\right)
    \quad\text{and}\quad
    b := \sum_{i\in C} w_i\frac{t_i}{\taus} \exp\left(\frac{t_i}{\taus}\right)
    \;.
\end{align}
These definitions use the causal set $C=\{i \; |\;  t_i<T\}$ of input spikes before the output.
With those definitions, the spike time of a neuron with identical membrane and synaptic time constant $\taum=\taus$ is
(\cref{eq:equalTimeEquation} in the main text)
\begin{equation}
    \T  =  \taus \left\{
                \frac{b}{a_1} - \mathcal{W}\!\left[
                       -\frac{\gL\Vth}{a_1} \exp\left(\frac{b}{a_1}\right)
                \right]
            \right\}
        \; ,
\end{equation}
and for $\taum=2\taus$ 
(\cref{eq:doubleTimeEquation} in the main text)
\begin{equation}
    \T  =  2\taus\ln\! \left[
                \frac{2a_1}{a_2 + \sqrt{a_2^2 - 4a_1\gL\Vth}}
            \right]
        \; .
\end{equation}
The Lambert W function is defined as the solution $h=\mathcal W (z)$ to the equation $z = h\exp(h)$.

\subsection{Extending the formalism for multiple spikes}\label{sec:si_math_multi}
\begin{wrapfigure}[28]{l}{0.4\linewidth}
    \vspace{-0.5cm}
    \centering
        \includegraphics[]{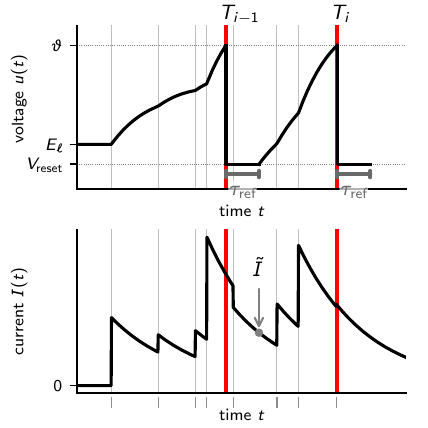}
    \caption{%
        \label{fig:SI_math_multiSpikeDynamics}
        \textbf{Sketch of the voltage (top) and current (bottom) dynamics of an \gls{lif} neuron that is spiking multiple times.}
        The gray vertical lines indicate the input spikes, the red vertical lines the output spikes.
        After each spike, the voltage is clamped to the reset potential $\Vreset$ for a time $\tauref$, highlighted by the gray vertical bar in the top plot.
        After this refractory period, the voltage evolves freely again, from $T_{i-1}+\tauref$ on driven by the residual current $\tilde I$ and additional input spikes.
    }
\end{wrapfigure}
In \cref{sec:theory}, we have derived our equations in a simplified scenario in which each neuron only spikes once.
As shown in the present and earlier work~\cite{goeltz2021fast}, single spikes can already be sufficient for many problems, however, for scenarios in which multiple spikes per neuron are necessary, the extended equations are derived below.

For the first spike of a neuron, \Cref{eq:equalTimeEquation,eq:doubleTimeEquation} are derived by integrating the \gls{ode} in \cref{eq:cubaLIF} to get voltage dynamics $u(t)$, and afterward solving for the time $T$ of the spike, defined by $u(T)=\Vth$.
Recalling the dynamics of the \gls{lif} model, after a spike the voltage is fixed at the reset voltage $\Vreset$ for a time $\tauref$ (\cref{fig:SI_math_multiSpikeDynamics}, top panel), and afterward continues to follow the dynamics laid out in the \gls{ode} (\ref{eq:cubaLIF}).
For the time of the second spike $T_2$ and all further spikes this implies that the same procedure can be followed when adhering to different initial conditions.

For the first spike, the integration of the \gls{ode} is performed with initial vanishing current $I(0)=0$ and voltage $u(0)=0$ (w.l.o.g.\ we have chosen leakage $\Vleak=0$), resulting in voltage dynamics
\begin{equation}
    \label{eq:SI_multi_voltage_simple}
    u(t) = \sum_{i \in C} \Theta(t-t_i) \frac{w_i}{\gl} \frac{\taus}{\taum-\taus} 
        \left[ \exp\left(-\frac{t - t_i}{\taum}\right) - \exp\left(-\frac{t - t_i} {\taus} \right) \right]
    \;.
\end{equation}
\linelabel{rev:refSImathMultiBC}
Assuming a spike at $T_{i-1}$, the new initial condition for the voltage can be written down at time $\tilde t := T_{i-1} + \tauref$ as
\begin{equation}
    \label{eq:SI_multi_IC}
    u(\tilde t) = \tilde u := \Vreset
    \;,
\end{equation}
while the current is not affected by the reset, and keeps following \cref{eq:Is}.
To simplify the notation, we split up the spikes of the causal set $t_i\in C$ into one set arriving before $\tilde t$,
    $C^<=\{t_i \in C | t_i < \tilde t\}$,
and one set with spikes afterward
    $C^\geq=\{t_i \in C | t_i \geq \tilde t\}$.
For times $t\geq \tilde t$, we can write the current as
\begin{equation}
    \label{eq:SI_multi_I}
    I(t) = \tilde I \exp\left(-\frac{t-\tilde t}{\taus} \right) + \sum_{i \in C^\ge} \Theta(t-t_i) w_i \exp\left( -\frac{t-t_i}{\taus} \right)
    \;,
\end{equation}
with $\tilde I = \sum_{i \in C^<} w_i \exp( -\frac{\tilde t-t_i}{\taus} )$ the (residual) current at time $\tilde t$ (see~\cref{fig:SI_math_multiSpikeDynamics}).
Integration of the \gls{ode} under these initial conditions results in an equation for the dynamics of the voltage:
\begin{equation}
    \begin{aligned}
    \label{eq:SI_multi_voltage}
    u(t) =
        &\sum_{i \in C^\geq} \Theta(t-t_i) \frac{w_i}{\gl} \frac{\taus}{\taum-\taus} 
            \left[ \exp\left(-\frac{t - t_i}{\taum}\right) - \exp\left(-\frac{t - t_i} {\taus} \right) \right]
      \\
        &+\tilde u \exp\left( -\frac{t - \tilde t}{\taum}\right)
        + \frac{\tilde I}{\gl} \frac{\taus}{\taum-\taus} 
            \left[ \exp\left(-\frac{t - \tilde t}{\taum}\right) - \exp\left(-\frac{t - \tilde t} {\taus} \right) \right]
    \;.
    \end{aligned}
\end{equation}
The first term encapsulates the effect of the incoming spikes as before (compare~\cref{eq:SI_multi_voltage_simple}),
while the second line is an exponential decay from the reset to the leak, while accounting for the effect of the residual current.
Interestingly, this form of the equation shows that the residual current can be modeled as a virtual spike with weight $\tilde I$ at time $\tilde t$.
Crucially, when starting at leak voltage $\tilde u=0$ and without initial current $\tilde I=0$, the above voltage dynamics \cref{eq:SI_multi_voltage_simple} are recovered.

With this equation, the derivation performed in~\cite{goeltz2021fast} can be followed, i.e., assuming a spike $T_i>\tilde t$ defines a causal set $\tilde C = \{t_i\in C^\geq | t_i < T_i\}$, and using $u(T_i) = \Vth$ we can write
\begin{equation}
    \begin{aligned}
    \label{eq:SI_multi_spikeTime}
    0 =
        &- \underbrace{ \left[
          \tilde I  \frac{\taus}{\taum - \taus} \exp\left( \frac{\tilde t }{\taus} \right)
          + \sum_{i \in \tilde C} w_i \frac{\taus} {\taum - \taus} \exp\left( \frac{t_i }{ \taus}\right)
      \right] 
          }_{a_1'}
      \cdot \exp\left(- \frac{T_i}{ \taus} \right)
      \\
        &+ \underbrace{  \left[
          \tilde u \exp\left( \frac{\tilde t }{ \taum}\right)
          + \tilde I  \frac{\taus}{\taum - \taus} \exp\left( \frac{\tilde t }{ \taum} \right)
          + \sum_{i \in \tilde C} w_i \frac{\taus} {\taum - \taus} \exp\left( \frac{t_i }{ \taum}\right)
      \right]
        }_{a_2'}
      \cdot \exp\left(- \frac{T_i }{ \taum} \right)
      \\
        &-\gl\Vth
    \\
   = & - a_1'  \exp\left(- \frac{T_i }{ \taus} \right)
   + a_2'  \exp\left(- \frac{T_i }{ \taum} \right)
        -\gl\Vth
      \;.
    \end{aligned}
\end{equation}
This equation follows the same structure as in the original derivation~\cite[Eq.~(26-27)]{goeltz2021fast} with differently defined parameters $a_1'$ and $a_2'$:
comparing with~\cref{eq:SI_mathAB}, in the case of $\taum=2\taus$ we find
\begin{equation}
    \begin{aligned}
    a_1
        &\rightarrow
        a_1' := a_1
            + \tilde I  \exp\left( \frac{\tilde t }{\taus} \right)
    \\
    a_2
        &\rightarrow
        a_2' := a_2
            + \tilde I   \exp\left( \frac{\tilde t }{ \taum} \right)
            + \tilde u \exp\left( \frac{\tilde t }{ \taum}\right)
    \;.
    \end{aligned}
\end{equation}
Notably, this implies that one can solve for $T_i$ to get a function
\begin{equation}
    \label{eq:SI_multi_T}
    T_i \left(\{t_i\} \cup \{w_i\}, \tilde I, \tilde t\right)
    \; .
\end{equation}
This function is at the heart of implementing exact, event-based forward dynamics and, more importantly, its differentiability enables error backpropagation through multiple layers of such neurons.
Through $\tilde I$ there is a dependence of the output spike times $T_i$ on earlier input spike times $t_i < \tilde t$, and through $\tilde t$ a dependence on the previous spike of the same neuron.
For the case of $\taum=\taus$, one can use l'Hôpital's rule and the Lambert W function to write down the solution for $T_i$.

\subsection{Explicit, event-based gradients of a voltage-max-over-time loss}\label{sec:si_math_vmax}
Typically, when using an event-based framework, information in a network is encoded in spike times.
For some tasks, losses based on the voltage of (a subset) of neurons have been proposed and used, especially the max-over-time loss (for a selection, see~\cite{cramer2022surrogate,goeltz2023gradients,bittar2022surrogate,wunderlich2021eventprop,nowotny2022loss}).
For this, the corresponding loss function depends on the correct class $n^\star$ and the voltage $\vect{u}(t)$ of the (non-spiking) label neurons together with a scale factor $a\_{scale}$ like
\begin{align}
    \label{eq:lossMOT}
        \mathcal{L}\_{ MOT}
        [\vect{u}(t)
        , n^\star; a\_{scale}]
        = -\log
        \Big[ \operatorname{softmax}_{n^\star} ( a\_{scale}\cdot\max\nolimits_t \vect{u}(t)) \Big]
        \; .
\end{align}
Because this loss has been predominantly used with surrogate gradients and depends on the membrane voltage of the label neurons, it is not typically associated with exact and event-based training schemes, with \cite{wunderlich2021eventprop,nowotny2022loss} being the exceptions.
However, the loss is compatible with a purely spike-based formulation and in the following the relevant gradient will be derived:
\begin{align}
    \frac{\partial u_{\max}}{\partial \theta} =
        \frac{\partial \shorthandU }{\partial\theta} \Big|_{t=\tilde t}
	+ \frac{\partial \shorthandU }{\partial t} \Big|_{t=\tilde t} \cdot \frac{\partial \tilde t}{\partial \theta}
	\;.
	\label{eq:SI_umax_first}
\end{align}
In addition to the natural first term, the second term occurs when an (inhibitory) input spike determines the maximum of the voltage.

For the derivation, we assume the voltage $u$ is a function of the parameters input spikes $\{t_i\}$ and weights $\{w_i\}$, here shortened as $\{\theta\}$
\begin{equation}
    u = u(t | \{t_i\} \cup \{w_i\})
     = u(t | \{\theta\})
    \;,
\end{equation}
the maximum voltage $u\_{max}$ at time $\tilde t$ is defined by
\begin{equation}
    \begin{array}{r  >{\displaystyle}c  l}
    \tilde t :=& \argmax_{t\in S} &u(t | \{\theta\})
    \\
        u\_{max} =& \max_{t\in S} &u(t | \{\theta\}) = u(\tilde t | \{\theta\})
    \;,
    \end{array}
\end{equation}
where $t\in\shorthandIntegrationDomain$ runs in the integration domain $\shorthandIntegrationDomain$.
This definition makes $u\_{max}$ an implicit function\footnote{In this section, we assume the maximum is uniquely defined in a neighborhood of the current parameters, i.e., there is no sudden jump of the maximum to another time.
    Cases in which this assumption is not satisfied have to be treated differently, e.g., by adding up gradients coming from these different maxima of equal value.
} of the parameters $\{\theta\}$.
Because the voltage of these (non-spiking) neurons is continuous, the maximization can be written in an integral form using the Dirac $\delta$ distribution
\begin{align}
    u\_{max} &= \max_t \shorthandU
        = u({\scriptstyle \tilde t | \{\theta\}})
           \\&= \int_\shorthandIntegrationDomain \intd t\, \shorthandU \delta(t - \tilde t)
        \label{eq:SI_math_integral}
           \;,
\end{align}
which will allow the reformulation in simple, event-based terms.

\begin{wrapfigure}[15]{r}{0.31\linewidth}
    \centering
    \input{figures//tikz/sketch_SI_math_umax/fig.tex}
    \caption{%
        \label{fig:SI_math_split_Uis}
        Separating $u$ into distinct, smooth $u_i$ at the input spike times $t_i$.
    }
\end{wrapfigure}
We split up the voltage along the spike times $t_i$, cf.~\cref{fig:SI_math_split_Uis}.
Here, w.l.o.g.\ we assume the input spikes $\{t_i|i\in[1, N]\}$ to be ordered $t_i<t_{i+1} \forall i$ and define $t_0=-\infty$ and $t_{N+1}=\infty$ as well as the intervals $S_i=(t_i, t_{i+1}]$.
Further, we want to define the shorthand $\tilde S$ for the interval $\tilde\imath$ that contains the maximum $ \tilde S := S_{\tilde \imath}  \ni \tilde t $,
therefore $t_{\tilde \imath}$ is the last input spike causal to the dynamics in this interval and there is no new spike within the interval.
Employing as a shorthand the synaptic interaction kernel 
\begin{equation}
    \kappa(t)=\Theta(t) \frac{\taus}{\taum-\taus} \left[\exp(-t/\taum) - \exp(-t/\taus)\right]
    \;,
\end{equation}
the voltage behaves like $\shorthandU = \frac{1}{\gl}  \sum_i^N w_i \kappa(t-t_i)$, and one can write

\begin{align}
    \shorthandU = \frac{1}{\gl} \cdot
    \begin{cases*}
        0 & if $t\le t_1$ \\
        w_1\kappa(t-t_1) & if $t_1<t\le t_2$ \\
        w_1\kappa(t-t_1) + w_2\kappa(t-t_2)& if $t_2<t\le t_3$ \\
        \qquad\vdots & $\quad\vdots$ \\
        \shorthandUi{n} & if $t_n<t\le t_{n+1}$
    \end{cases*}
    \;,
\end{align}
with $u_n(t|\{\theta\}) = \frac{1}{\gl} \sum_i^n w_i \kappa(t-t_i)$.
This function $u_n(t)$ is time-differentiable everywhere on its domain $S_n$.
The separation into $u_n$ can be carried over to all integrals of the voltage with any function $f$
\begin{align}
    \int_\shorthandIntegrationDomain \intd t\, \shorthandU \big[f(t)\big]
    = \sum_i \int_{S_i} \intd t\, \shorthandUi{i} \big[f(t)\big]
    \;.
    \label{eq:SI_umax_sum_of_integral}
\end{align}

Specifically, this can be done for the integral formulation of $u\_{max}$~\cref{eq:SI_math_integral}.
For the next step, a requirement is the derivative of an integral with varying boundaries:
\begin{align}
    \frac{\partial}{\partial y}\int_{a(y)}^{b(y)}\intd x\, f(x, y) = \biggl[ f(x, y) \frac{\partial}{\partial y} x \biggr]_{x=a(y)}^{x=b(y)} + \int_{a(y)}^{b(y)}\intd x\, \frac{\partial}{\partial y} f(x, y)
    \,.
    \label{eq:SI_umax_varying_boundaries}
\end{align}

Differentiating the integral form of the maximum voltage leads to:
\begin{align}
    \frac{\partial u_{\max}}{\partial \theta}
    &= \frac{\partial }{\partial \theta} \int_\shorthandIntegrationDomain \intd t\, \shorthandU \cdot \delta(t - \tilde t)
\\  &= \frac{\partial }{\partial \theta} \sum_i \underbrace{\int_{S_i} \intd t\, \shorthandUi{i} \cdot \delta(t - \tilde t)}_{\text{vanishes due to $\delta$ except if $i=\tilde \imath$}}
\\  &= \frac{\partial }{\partial \theta} \int_{t_{\tilde \imath}}^{t_{\tilde \imath+1}} \intd t\, \shorthandUi{\tilde \imath} \cdot \delta(t - \tilde t)
    \\
    &= \biggl[\shorthandUi{\tilde\imath} \cdot \delta(t - \tilde t) \frac{\partial t}{\partial \theta}\biggr]_{t_{\tilde \imath}}^{t_{\tilde \imath + 1}}
    + \int_{t_{\tilde \imath}}^{t_{\tilde \imath+1}} \intd t\,
    \frac{\partial}{\partial\theta} \left[ \shorthandUi{\tilde\imath} \cdot \delta(t - \tilde t)  \right]
  \\&= \biggl[\shorthandUi{\tilde\imath} \cdot \delta(t - \tilde t) \frac{\partial t}{\partial \theta} \biggr]_{t_{\tilde \imath}}^{t_{\tilde \imath + 1}}
    +\int_{t_{\tilde \imath}}^{t_{\tilde \imath+1}} \intd t\,
        \left[
	\frac{\textstyle \partial \shorthandUi{\tilde\imath}}{\textstyle \partial \theta}  \cdot \delta(t - \tilde t)
    + \shorthandUi{\tilde\imath} \cdot \frac{\partial }{\partial \theta}  \delta(t - \tilde t)
        \right]
  \\&= \biggl[\shorthandUi{\tilde\imath} \cdot \delta(t - \tilde t) \frac{\partial t}{\partial \theta} \biggr]_{t_{\tilde \imath}}^{t_{\tilde \imath + 1}}
    + \frac{\partial \shorthandU}{\partial\theta}\big|_{t=\tilde t}
    + \int_{t_{\tilde \imath}}^{t_{\tilde \imath+1}} \intd t\,
    \shorthandUi{\tilde\imath} \cdot\frac{\partial }{\partial \theta}  \delta(t - \tilde t)
	\label{eq:SI_umax_end}
	\;.
\end{align}

For the last term, the derivative of the delta distribution $\frac{\partial }{\partial \theta}  \delta(t - \tilde t)$ can be reformulated:
\begin{align}
     \frac{\partial  \delta(t - \tilde t)}{\partial \theta}
    &= \frac{\partial \delta (t - \tilde t)}{\partial (t - \tilde t)} \frac{\partial (t - \tilde t)}{\partial \theta}
  \\&= -\frac{\partial \delta (t - \tilde t)}{\partial (t - \tilde t)} \frac{\partial \tilde t}{\partial \theta}
  \\&= - \frac{\partial \delta (t - \tilde t)}{\partial (t - \tilde t)} \overbrace{ \frac{\partial (t - \tilde t)}{\partial t}}^{\text{inserting }1} \frac{\partial \tilde t}{\partial \theta}
    \\&= - \frac{\partial \delta (t - \tilde t)}{\partial t} \frac{\partial \tilde t}{\partial \theta}
    \;.
    \label{eq:SI_umax_star}
\end{align}

Inserting this above in~\cref{eq:SI_umax_end} and integrating by parts\footnote{
    $\int\intd x\, f(x) \frac{\partial g(x)}{\partial x} = f(x) g(x) \big|\_{boundary} - \int\intd x\, \frac{\partial f(x)}{\partial x} g(x)$
} allows removing the derivative of the $\delta$ distribution:

\begin{align}
    \int_{t_{\tilde \imath}}^{t_{\tilde \imath+1}} \intd t\,
    \shorthandUi{\tilde\imath} \cdot\frac{\partial }{\partial \theta}  \delta(t - \tilde t)
	&=-\int_{t_{\tilde \imath}}^{t_{\tilde \imath+1}} \intd t\,
		\shorthandUi{\tilde\imath}
		\frac{\partial \delta (t - \tilde t)}{\partial t}
        \frac{\partial \tilde t}{\partial \theta}
    \\&=-\left[ \shorthandUi{\tilde\imath} \delta(t-\tilde t) \frac{\partial \tilde t}{\partial \theta} \right]_{t_{\tilde \imath}}^{t_{\tilde \imath + 1}}
	    + \int_{t_{\tilde \imath}}^{t_{\tilde \imath+1}} \intd t\,
		\delta (t - \tilde t) \frac{\partial}{\partial t} 
        \Big(\shorthandUi{\tilde\imath} \cdot \frac{\partial \tilde t}{\partial \theta} \Big)
    \\&=-\left[ \shorthandUi{\tilde\imath} \delta(t-\tilde t) \frac{\partial \tilde t}{\partial \theta} \right]_{t_{\tilde \imath}}^{t_{\tilde \imath + 1}}
	    + \int_{t_{\tilde \imath}}^{t_{\tilde \imath+1}} \intd t\,
		\delta (t - \tilde t) \cdot
        \frac{\partial \shorthandUi{\tilde\imath} }{\partial t} \cdot \frac{\partial \tilde t}{\partial \theta}
    \\&=-\left[\shorthandUi{\tilde\imath} \delta(t - \tilde t) \frac{\partial \tilde t}{\partial \theta} \right]_{t_{\tilde \imath}}^{t_{\tilde \imath + 1}}
        + \frac{\partial \shorthandUi{\tilde\imath} }{\partial t} \Big|_{t=\tilde t} \cdot \frac{\partial \tilde t}{\partial \theta}
    \\&=-\left[\shorthandUi{\tilde\imath} \delta(t - \tilde t) \frac{\partial \tilde t}{\partial \theta} \right]_{t_{\tilde \imath}}^{t_{\tilde \imath + 1}}
        + \frac{\partial \shorthandU }{\partial t} \Big|_{t=\tilde t} \cdot \frac{\partial \tilde t}{\partial \theta}
        \;.
    \label{eq:SI_umax_star_reformulation}
\end{align}
The time derivative only acts on $u_{\tilde\imath}$ because $\tilde t$ (and its derivative) are independent of the integration variable $t$.
Furthermore, at time $\tilde t$ the value of $u_{\tilde\imath}$ is identical to the one of $u$, so we can substitute the regular voltage back in.

With \cref{eq:SI_umax_star_reformulation} inserted into \cref{eq:SI_umax_end}, reordering of the terms yields 
\begin{align}
    \frac{\partial u_{\max}}{\partial \theta} 
    &=\biggl[\shorthandU \cdot \delta(t - \tilde t) \frac{\partial t}{\partial \theta} \biggr]_{t_{\tilde \imath}}^{t_{\tilde \imath + 1}}
    -\biggl[\shorthandU \cdot \delta(t - \tilde t) \frac{\partial \tilde t}{\partial \theta} \biggr]_{t_{\tilde \imath}}^{t_{\tilde \imath + 1}}
    + \frac{\partial \shorthandU }{\partial \theta} \Big|_{t=\tilde t}
    + \frac{\partial \shorthandU }{\partial t} \Big|_{t=\tilde t} \frac{\partial \tilde t}{\partial \theta}
    \;.
    \label{eq:SI_umax_long_summary}
\end{align}

From $\tilde t \in S_{\tilde \imath}=(t_{\tilde \imath}, t_{\tilde \imath+1}]$ follows $\tilde t \neq t_{\tilde \imath}$, therefore the first two terms evaluated at the lower boundary vanish.
The remaining, upper boundary terms $t=t_{\tilde\imath+1}$ are only nonzero if the maximum happens at that boundary $\tilde t = t_{\tilde \imath + 1}$, in which case the two terms cancel each other, yielding the final, concise result
\begin{align}
    \frac{\partial u_{\max}}{\partial \theta} =
        \frac{\partial \shorthandU }{\partial\theta} \Big|_{t=\tilde t}
	+ \frac{\partial \shorthandU }{\partial t} \Big|_{t=\tilde t} \cdot \frac{\partial \tilde t}{\partial \theta}
	\;.
	\label{eq:SI_umax_final}
\end{align}

There are two distinct cases how a maximum of the voltage can be reached:
The more common one is a maximum due to the decay of the voltage back to the leakage.
In a neighborhood around this maximum at $\tilde t$, the voltage is a smooth function with time derivative $\frac{\partial u}{\partial t}|_{t=\tilde t} = \dot u(\tilde t) = 0$.
Therefore, the second term in~\cref{eq:SI_umax_final} vanishes.

The other possibility is in the event of a sufficiently strong inhibitory spike:
as a consequence, the voltage can decrease immediately and the time of maximal voltage is identical to the time of this inhibitory input $\tilde t =t_{\tilde \imath + 1}$.
In this case, the second term will be nonzero but can be calculated because both $\dot u$ and $\frac{\partial \tilde t}{\partial \theta} = \frac{\partial t_{\tilde \imath + 1}}{\partial \theta}$ of the inhibitory input spike $t_{\tilde\imath +1}$ are known.
This contribution is proportional to $\dot u$ (the left derivative at time of input spike, i.e., how much the membrane changes in free dynamics) as well as $\Delta t$ (how much a change in parameter $\theta$ influences the relevant input spike time $t_{\tilde\imath +1}$).

Now, we investigate
$
    \frac{\partial u(t)}{\partial \dots} \Bigr|_{t=\tilde t}
$
in two different settings, starting with the more peculiar one.

\paragraph{Equal time constants $\boldsymbol{\taus=\taum}$ }
In this regime the voltage behaves as
\begin{align}
    u(t) &= \sum_i \Theta(t-t_i) \frac{w_i}{\gl}  \frac{t-t_i}{\taus} \exp\left(-\frac{t-t_i}{\taus}\right)
    \; .
\end{align}
We can calculate the derivative to be
\begin{align}
    \frac{\partial u(t)}{\partial w_j} \Bigr|_{t=\tilde t}
        &= \frac1{\gl}  \sum_{i\in\{i|t_i < \tilde t \} }
            \underbrace{\frac{\partial w_i}{\partial w_j} }_{\delta_{ij}}
            \frac{\tilde t - t_i}{\taus} \exp\left(-\frac{\tilde t - t_i}{\taus}\right)
    \\
        &= \frac1{\gl}  \mathbb{1}_{t_j < \tilde t }
            \frac{\tilde t - t_j}{\taus} \exp\left(-\frac{\tilde t - t_j}{\taus}\right)
        \; .
\end{align}
Similarly, we get
\begin{align}
    \frac{\partial u(t)}{\partial t_j} \Bigr|_{t=\tilde t}
        &= \frac{w_j}{\gl}  \mathbb{1}_{t_j < \tilde t }
            \frac{\tilde t - t_j - \taus}{\taus^2}
            \exp\left(-\frac{\tilde t - t_j}{\taus}\right)
            \; .
\end{align}

\paragraph{Unmatched time constants $\boldsymbol{\taus\neq\taum}$ }
While the voltage dynamics is slightly different
\begin{align}\label{eq:si_umem_tauDifferent}
    u(t) &=
        \sum_i \Theta(t-t_i) \frac{w_i}{\gl}
        \frac{\taus}{\taum - \taus}
        \left[
            \exp\left(-\frac{t-t_i}{\taum}\right) - \exp\left(-\frac{t-t_i}{\taus}\right)
        \right]
        \; ,
\end{align}
the calculation is similar and results in
\begin{align}
    \frac{\partial u(t)}{\partial w_j} \Bigr|_{t=\tilde t}
        &= \frac{\taus}{\taum - \taus} \frac1{\gl}
            \mathbb{1}_{t_j < \tilde t }
            \left[
                \exp\left(-\frac{\tilde t-t_j}{\taum}\right) - \exp\left(-\frac{\tilde t-t_j}{\taus}\right)
            \right]
    \\
    \frac{\partial u(t)}{\partial t_j} \Bigr|_{t=\tilde t}
        &= \frac{\taus}{\taum - \taus}
            \frac{w_j}{\gl} \mathbb{1}_{t_j < \tilde t }
            \left[
                 \frac1\taum \cdot \exp\left(-\frac{\tilde t-t_j}{\taum}\right)
                - \frac1\taus \cdot \exp\left(-\frac{\tilde t-t_j}{\taus}\right)
            \right]
            \; .
\end{align}

\subsection{Relationship of weight and delay for LIF-based parrot neuron}\label{sec:si_lif_delay_theoretical}

With the \gls{lif} dynamics from above (\cref{eq:si_umem_tauDifferent}), we can proceed to get the desired relationship of a weight and the resulting delay.
To calculate the delay of a parrot neuron that has one input spike time $t$ associated with a weight $w$, one needs to compute its time of spiking (i.e., $u(T)=\vartheta$).
Assuming w.l.o.g.\ $t=0$ and using $T=t+d$ for a delay $d$ yields
\begin{align}
\vartheta= \frac{\tau_{\mathrm{s}}}{g_{\ell}(\tau_{\mathrm{m}}-\tau_{\mathrm{s}})} w \left[\exp \left(-\frac{d}{\tau_{\mathrm{m}}}\right) - 
 \exp \left(-\frac{d}{\tau_{\mathrm{s}}}\right) \right]
 \;.
\end{align}
Solving for the weight $w$ returns
\begin{align}
w = \frac{g_{\ell} \vartheta (\tau_{\mathrm{m}}-\tau_{\mathrm{s}})} { \tau_{\mathrm{s}} } \frac{1}{ \exp \left(-\frac{d}{\tau_{\mathrm{m}}}\right) - 
 \exp \left(-\frac{d}{\tau_{\mathrm{s}}}\right) }
 \;.
 \label{eq:SI_wofd}
\end{align}

However, this holds only if the neuron is in fact spiking.
This can happen in the interval $[0;\tilde{t}]$ with $\tilde{t}$ the time at which the membrane voltage is maximal:
\begin{align}
    \tilde{t} &= \frac{ \taum\taus }{ \taum - \taus} \log \frac{\taum}{\taus}
    \;.
 \label{eq:SI_ttilde}
\end{align}

For the specific case of $\tau_{\mathrm{s}} = \tau_{\mathrm{m}}$, l'Hôpital's rule in the limit $\tau_{\mathrm{m}} \rightarrow \tau_{\mathrm{s}}$ can be applied to \cref{eq:SI_wofd} and \cref{eq:SI_ttilde}:
\begin{align}
    w = \frac{g_{\ell} \vartheta \tau_{\mathrm{s}}}{d} \exp \left(\frac{d}{ \tau_{\mathrm{s}}}\right)
    \quad
    \text{and}
    \quad
    \tilde{t} =  \tau_{\mathrm{s}}
    \;.
\end{align}

\subsection{Simulation parameters}
\label{sec:SI_simulation_parameters}

\begin{table}
    \centering
    \caption{\textbf{Dataset and training parameters}. Used to produce the results in  \cref{fig:simulation}, \cref{fig:hw_analog_parrot}, \cref{fig:deeper_networks},  \cref{fig:SI_extended_results}, \cref{fig:SI_extended_results_randButFixed}, \cref{fig:si_hw_aware_sim}, \cref{fig:SI_yy_comparison} and \cref{tab:hw_aware_sim}.}
    \begin{threeparttable}[!ht]
    \centering
    \begin{tabular}{l | >{\centering\arraybackslash}m{5cm} >{\centering\arraybackslash}m{5cm}}
    \textbf{parameter name} & \textbf{ideal simulation} & \textbf{hardware-aware simulation/\newline hardware emulation} \\
    \hline
    \textbf{dataset parameters} & & \\
    input size & 4 & 4 \\
    $t_{\text{early}}$ & 0.15 & 0.15 \\
    $t_{\text{late}}$ & 2.0 & 2.0 \\
    \hline
    \textbf{training parameters} & & \\
    training epochs & 300 & 300 \\
    batch size & 150 & 40 \\
    adam parameter $\beta$ & $(0.9,0.999)$ & $(0.9,0.999)$ \\
    adam parameter $\epsilon$ & $10^{-8}$ & $10^{-8}$ \\
    lr-scheduler & StepLR & StepLR \\
    lr-scheduler step size & 20 & 20 \\
    lr-scheduler $\gamma$ & 0.95 & 0.95 \\
    delay-lr\tnote{1} & $[0.1, 0.3, 0.5, 1, 1.5, 2] \times 10^{-2}$ & $2 \times 10^{-3}$ \\
    weight-lr\tnote{1} & $[0.1, 0.3, 0.5, 1, 1.5, 2] \times 10^{-2}$ & $2 \times 10^{-3}$ \\
    input noise $\sigma$ & no noise & no noise \\
    max allowed $\Delta w$ & 0.2 & 0.2 \\
    weight bump value & 0.0005 & 0.0005 \\
    loss $\Delta_t$ & 0.2 & 0.3 \\
    \end{tabular}
        \begin{tablenotes}
            \footnotesize
            \item[1] For hyperparameter optimization, a grid search was performed over the range of values in brackets.
        \end{tablenotes}
        \label{tab:SI_HP_parameters_training}
    \end{threeparttable}
\end{table}

\begin{table}
    \centering
    \caption{\textbf{Network parameters.} Used to produce the results in \cref{fig:simulation}, \cref{fig:hw_analog_parrot}, \cref{fig:deeper_networks},  \cref{fig:SI_extended_results}, \cref{fig:SI_extended_results_randButFixed}, \cref{fig:si_hw_aware_sim}, \cref{fig:SI_yy_comparison} and \cref{tab:hw_aware_sim}.
    }
    \begin{threeparttable}[h]
        \centering
        
    \begin{tabular}{l | >{\centering\arraybackslash}m{5cm} >{\centering\arraybackslash}m{5cm}}
    \textbf{parameter name} & \textbf{ideal simulation} & \textbf{hardware-aware simulation/\newline hardware emulation} \\
    \hline
    \textbf{neuron parameters} & & \\
    $g_{\ell}$ & 0.5 & 1.0 \\
    $E_{\ell}$ & 0.0 & 0.0 \\
    $\vartheta$ & 1.0 & 2.6 \\
    $\tau_{\mathrm{m}}$ & 2.0 & 1.0 \\
    $\tau_{\mathrm{s}}$ & 1.0 & 1.0 \\
    \hline
    \textbf{network parameters} & & \\
    \textit{layer 0\tnote{1}} & \multicolumn{2}{c}{[\textit{broadcast, axonal, dendritic, synaptic}]} \\
    delay init mean & 0.0 & 0.0 \\
    delay init std & 0.25 & 0.5 \\
    scale $\lambda$ & 1.0 & 1.5 \\
    shift & 0.0 & 2.0 \\
    \\
    \textit{layer 1} & \multicolumn{2}{c}{\textit{neuron}} \\
    size\tnote{1} &  \multicolumn{2}{c}{[5, 10, 15, 20, 25, 30]}\\
    max ratio missing spikes & 0.3 & 0.05 \\
    weight init mean & 1.0 & 1.0 \\
    weight init std & 1.0 & 0.12 \\
    \\
    \textit{layer 2\tnote{1}} & \multicolumn{2}{c}{[\textit{broadcast, axonal, dendritic, synaptic}]} \\
    delay init mean & 0.0 & 0.0 \\
    delay init std & 0.25 & 0.5 \\
    scale $\lambda$ & 1.0 & 1.5 \\
    shift & 0.0 & 2.0 \\
    \\
    \textit{layer 3} & \multicolumn{2}{c}{\textit{neuron}} \\
    size & 3 & 3 \\
    max ratio missing spikes & 0.0 & 0.05 \\
    weight init mean & 1.0 & 0.075 \\
    weight init std & 1.0 & 0.15 \\
    \end{tabular}
        \begin{tablenotes}
            \footnotesize
            \item[1] Parameters over which a sweep was performed are in brackets.
        \end{tablenotes}
        \label{tab:SI_HP_parameters_networks}
    \end{threeparttable}
\end{table}
\newpage

\begin{table}
    \centering
    \caption{%
        \textbf{Weight and delay configurations.} Used to produce the results in \cref{fig:SI_extended_results}.
        These values are extracted from trained networks and averaged from $10$ different seeds.
        They were not trained after the initialization, but rather fixed.
        The rest of the parameters were kept the same as in \cref{tab:SI_HP_parameters_networks}.
        }
    \begin{threeparttable}[h]
        \centering
        \begin{tabular}{l|cccc}
            \textbf{parameter name} & \multicolumn{4}{c}{\textbf{ablation study}} \\
            \hline
            \\
            \textit{layer 0} & \textit{broadcast} & \textit{axonal} & \textit{dendritic} & \textit{synaptic} \\
            delay mean & & 0.0 & 0.0 & 0.0 \\
            delay std & & 0.50 & 1.13 & 0.92 \\
            \\
            \textit{layer 1} & \multicolumn{4}{c}{\textit{neuron}} \\
            weight mean & 0.68 & 0.87 & 0.91 & 0.90 \\
            weight std & 1.14 & 1.72 & 1.47 & 1.17 \\
            \\
            \textit{layer 2} & \textit{broadcast} & \textit{axonal} & \textit{dendritic} & \textit{synaptic} \\
            delay mean & & 0.0 & 0.0 & 0.0 \\
            delay std & & 0.85 & 0.21 & 0.75 \\
            \\
            \textit{layer 3} & \multicolumn{4}{c}{\textit{neuron}} \\
            weight mean & 0.56 & 1.71 & 1.64 & 1.26 \\
            weight std & 1.53 & 4.55 & 3.88 & 2.12 \\
    
        \end{tabular}

    \end{threeparttable}
\end{table}
\newpage

\begin{table}
    \centering
    \caption{
    \textbf{Specific delay configurations.} Used to produce the results in \cref{fig:SI_extended_results_randButFixed}.
    Those values were not trained after the initialization, but rather fixed.
    The rest of the parameters were kept the same as in \cref{tab:SI_HP_parameters_networks}.
    }
    \begin{threeparttable}[h]
        \centering
    \begin{tabular}{l|c}
    \textbf{parameter name} & \textbf{random but fixed delay} \\
    \hline
    \\
    \textit{Layer 0} & \textit{broadcast, axonal, dendritic, synaptic} \\
    delay mean & 0.0 \\
    delay std\tnote{1} & [0.0, 0.4375, 0.875, 1.3125, 1.75] \\
    \\
    \textit{Layer 2} & \textit{broadcast, axonal, dendritic, synaptic} \\
    delay mean & 0.0 \\
    delay std\tnote{1} & [0.0, 0.4375, 0.875, 1.3125, 1.75]\\
    \end{tabular}
        \begin{tablenotes}
            \footnotesize
            \item[1] For hyperparameter optimization, a grid search was performed over the range of values in brackets.
        \end{tablenotes}
    \end{threeparttable}
\end{table}

\begin{table}
    \centering
    \caption{%
        \label{tab:SI_hw_aware_values}
        \textbf{Hardware parameters.} Used to produce the results in \cref{fig:hw_analog_parrot}, \cref{fig:si_hw_aware_sim}, \cref{fig:SI_yy_comparison} and \cref{tab:hw_aware_sim}.
        These values were obtained from the study made in \cref{sec:si_hardware_aware_sim}.
    }
    \begin{threeparttable}[h]
        \begin{tabular}{l|c}
        \textbf{parameter name} & \textbf{hw-aware} \\
        \hline
        \\
        weight quant. (max range)\tnote{1} & 2.1 \\
        weight quant. (precision) & 1/30 \\
        FP noise (mean) & 0.13 \\
        FP noise (std) & 0.08 \\
        trial-to-trial (std) & 0.04 \\
        delay jitter & 0.01 \\
        \end{tabular}
        \begin{tablenotes}
            \footnotesize
            \item[1] To ensure sufficient drive at the input, the maximum range of the weight was multiplied by $5$ for networks with a hidden layer of $5$ neurons, for both hardware-aware simulation and hardware emulation.
        \end{tablenotes}
    \end{threeparttable}
\end{table}

\end{document}

%% file: glossaries.tex
\newacronym{adc}{ADC}{analog-to-digital converter}
\newacronym{adex}{AdEx}{adaptive exponential leaky integrate-and-fire}
\newacronym{ai}{AI}{Artificial Intelligence}
\newacronym{ann}{ANN}{artificial neural network}
\newacronym{bss2}{BSS-2}{BrainScaleS-2}
\newacronym{cmos}{CMOS}{Complementary Metal-Oxide-Semiconductor}
\newacronym{fnd}{FnD}{Fast\&Deep}
\newacronym{fpga}{FPGA}{field-programmable gate array}
\newacronym{hpo}{HPO}{Hyperparameter Optimization}
\newacronym{iqr}{IQR}{interquartile range}
\newacronym{lif}{LIF}{leaky integrate-and-fire}
\newacronym{mse}{MSE}{mean squared error}
\newacronym{nlif}{nLIF}{non-leaky integrate-and-fire}
\newacronym{ode}{ODE}{ordinary differential equation}
\newacronym{psp}{PSP}{postsynaptic potential}
\newacronym{rnn}{RNN}{recurrent neural network}
\newacronym{rram}{RRAM}{Resistive Random Access Memory}
\newacronym{snn}{SNN}{spiking neural network}
\newacronym{srm}{SRM}{spike response model}
\newacronym{stdp}{STDP}{spike-timing-dependent plasticity}
\newacronym{ttfs}{TTFS}{time-to-first-spike}
\newacronym{yy}{YY}{Yin-Yang}

%% file: table_hw_aware.tex
\begin{table}[h!]
    \caption{%
        \textbf{Estimation of the impact of different hardware phenomena on training results using the hardware-aware simulation framework.}
        The values given are the median test errors with the \gls{iqr} in parentheses.
    }
    \centering
    \small
    \sisetup{parse-numbers=false}
    \resizebox{\textwidth}{!}{\begin{tabular}{p{2.2cm}p{2.3cm}p{2.3cm}p{2.3cm}p{2.3cm}p{2.3cm}p{2.3cm}p{2.3cm}}
    \toprule
         & ideal simulation & simulation \newline(HW params) & simulation \newline(HW params) \newline + weight quant. & simulation \newline(HW params) \newline + weight quant. \newline + FP noise & simulation \newline(HW params) \newline + weight quant. \newline + FP noise \newline + trial-to-trial & simulation \newline(HW params) \newline + weight quant. \newline + FP noise \newline + trial-to-trial \newline + delay jitter & hardware \\ \midrule
        weights\newline + axonal delays  & \SI{2.60^{4.67}_{2.15}}{\percent} & \SI{4.15^{5.55}_{3.55}}{\percent} & \SI{5.05^{7.03}_{4.30}}{\percent} & \SI{5.60^{9.03}_{4.30}}{\percent} & \SI{7.90^{8.75}_{7.10}}{\percent} & \SI{9.30^{10.52}_{7.83}}{\percent} & \SI{7.40^{9.35}_{6.93}}{\percent} \\
        \midrule
        weights  & \SI{3.20^{3.80}_{2.48}}{\percent} & \SI{8.15^{8.70}_{7.00}}{\percent} & \SI{9.60^{12.27}_{7.80}}{\percent} & \SI{9.70^{10.42}_{8.33}}{\percent}&\multicolumn{2}{c}{\SI{12.50^{13.75}_{10.97}}{\percent}}  & \SI{13.95^{15.34}_{11.97}}{\percent} \\
    \bottomrule
    \end{tabular}}\label{tab:hw_aware_sim}
\end{table}

%% file: figures/tikz/sketch_SI_math_umax/fig.tex
\begin{tikzpicture}[scale=0.75]
    \def\arrSpiketimes{{1, 2, 3, 4}}
    \def\spiketimeA{1}
    \def\spiketimeB{2}
    \def\spiketimeC{3}
    \def\spiketimeD{4}
    \def\weightA{0.6}
    \def\weightB{-0.3}
    \def\weightC{0.4}
    \def\weightD{0.7}

    \def\heightLabelUi{0.5}
    \def\valueXlim{5}
    
    \def\arrSpiketimes{{\spiketimeA, \spiketimeB, \spiketimeC, \spiketimeD}}

    \pgfmathdeclarefunction{Heaviside}{1}{%
      \pgfmathparse{#1<0 ? 0 : 1}%
    }

    \begin{axis}[
        xlabel={$t$},
        ylabel={$u_\text{mem}$},
        domain=0:\valueXlim,
        samples=100,
        axis lines*=middle,
        ymin=0, ymax=0.65,
	xmin=0, xmax=\valueXlim,
        xtick={\arrSpiketimes},
        xticklabels={$t_1$, $t_2$, $t_3$, $t_N$},
        ytick={0},
        yticklabels={},
        clip=false,
    ]

        \addplot[
            samples=200,
            thick,
        ] {
            Heaviside(x-\spiketimeA)*(x-\spiketimeA)*\weightA*exp(-(x-\spiketimeA))+
            Heaviside(x-\spiketimeB)*(x-\spiketimeB)*\weightB*exp(-(x-\spiketimeB))+
            Heaviside(x-\spiketimeC)*(x-\spiketimeC)*\weightC*exp(-(x-\spiketimeC))+
            Heaviside(x-\spiketimeD)*(x-\spiketimeD)*\weightD*exp(-(x-\spiketimeD))
        };
        
        \foreach \i in {0,...,3} {
            \pgfmathsetmacro\xval{\arrSpiketimes[\i]}
            \addplot[
                domain=0:\heightLabelUi,
                samples=10,
                dashed,
                black,
            ] ({\xval}, {x});
        }
        \coordinate (origin) at (0, 0);
        \coordinate (maxright) at (\valueXlim, 0);

        \coordinate (line-A) at (\spiketimeA, \heightLabelUi);
        \coordinate (line-B) at (\spiketimeB, \heightLabelUi);
        \coordinate (line-C) at (\spiketimeC, \heightLabelUi);
        \coordinate (line-D) at (\spiketimeD, \heightLabelUi);

        \node[align=right] at ([xshift=-15.3]line-A) {$u_0$};
        \node at ($(line-A)!0.5!(line-B)$) {$u_1$};
        \node at ($(line-B)!0.5!(line-C)$) {$u_2$};
        \node at ($(line-C)!0.5!(line-D)$) {$\cdots$};
        \node[align=left] at ([xshift=15.3]line-D) {$u_N$};

        \draw[|-|] (\spiketimeA, -0.06) -- node[below] {$S_1$}  (\spiketimeB, -0.06);

    \end{axis}
\end{tikzpicture}